\documentclass[10pt,twocolumn,letterpaper]{article}

\usepackage[pagenumbers]{cvpr} 

\usepackage{graphicx}
\usepackage[table]{xcolor}   
\usepackage{colortbl}        
\usepackage{array,booktabs}  
\usepackage{caption}         

\usepackage{booktabs}
\usepackage{makecell}
\usepackage[table]{xcolor}
\usepackage{graphicx}
\definecolor{navyblue}{RGB}{0,86,125}

\newcommand{\bench}{VSI-Bench}

\usepackage{multirow}
\definecolor{navyblue}{HTML}{0071BC}
\definecolor{hotpink}{HTML}{FF0080}

\definecolor{oai-white}{HTML}{FFFFFF}
\definecolor{oai-black}{HTML}{000000}
\definecolor{oai-red}{HTML}{FF4500}
\definecolor{oai-green}{HTML}{51DA4C}
\definecolor{oai-blue}{HTML}{0000FF}
\definecolor{oai-yellow}{HTML}{FFF639}
\definecolor{oai-magenta}{HTML}{FF45FF}
\definecolor{oai-cyan}{HTML}{00FFFF}
\definecolor{oai-orange}{HTML}{FE7600}
\definecolor{oai-violet}{HTML}{8A2BE2}
\definecolor{oai-brown}{HTML}{A0522D}
\definecolor{oai-green-050}{HTML}{F4FFF4}
\definecolor{oai-green-100}{HTML}{E9FFE8}
\definecolor{oai-green-200}{HTML}{D9FFD8}
\definecolor{oai-green-300}{HTML}{C9FFC7}
\definecolor{oai-green-400}{HTML}{A6FFA3}
\definecolor{oai-green-500}{HTML}{7CF178}
\definecolor{oai-green-600}{HTML}{51DA4C}
\definecolor{oai-green-700}{HTML}{3FA93B}
\definecolor{oai-green-800}{HTML}{2D712A}
\definecolor{oai-green-900}{HTML}{193718}
\definecolor{oai-gray-000}{HTML}{FFFFFF}
\definecolor{oai-gray-100}{HTML}{FAFAFA}
\definecolor{oai-gray-200}{HTML}{F5F5F5}
\definecolor{oai-gray-300}{HTML}{E5E5E5}
\definecolor{oai-gray-400}{HTML}{FFB7A4}
\definecolor{oai-gray-500}{HTML}{CDCDCD}
\definecolor{oai-gray-600}{HTML}{A8A8A8}
\definecolor{oai-gray-700}{HTML}{747474}
\definecolor{oai-gray-800}{HTML}{393939}
\definecolor{oai-gray-900}{HTML}{000000}

\definecolor{cvprblue}{rgb}{0.21,0.49,0.74}
\usepackage[pagebackref,breaklinks,colorlinks,allcolors=cvprblue]{hyperref}

\def\paperID{17342} 
\def\confName{CVPR}
\def\confYear{2026}

\title{SpatialStack: Layered Geometry-Language Fusion \\ for 3D VLM Spatial Reasoning}

\author{
Jian Zhang$^1$\thanks{Equal contribution.} \quad
Shijie Zhou$^{2,3}$\footnotemark[1]\quad
Bangya Liu$^4$\footnotemark[1] \quad
Achuta Kadambi$^2$ \quad  
Zhiwen Fan$^5$\\[0.4em]
\normalsize{
$^1$XMU \quad
$^2$UCLA \quad
$^3$Google\quad
$^4$UW-Madison\quad
$^5$TAMU
}\\[0.3em]
\url{https://spatial-stack.github.io/}
}

\begin{document}

\twocolumn[{%
\renewcommand\twocolumn[1][]{#1}%
\maketitle
    \captionsetup{type=figure}
    \vspace{-8mm}
    \includegraphics[width=\textwidth]{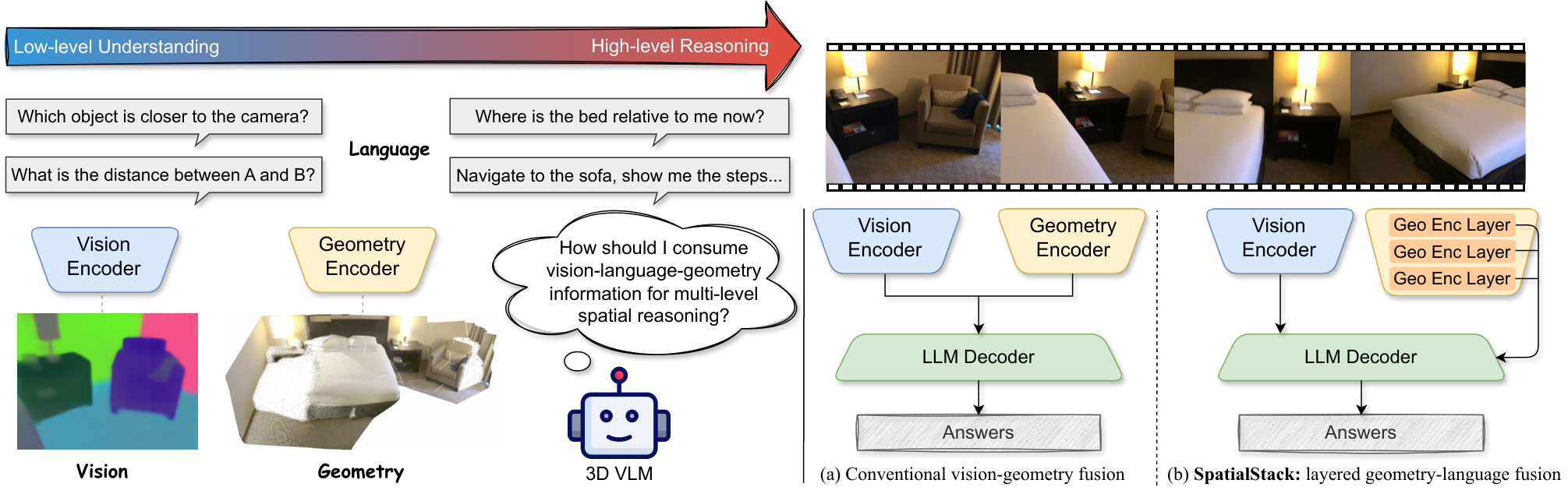}
    \vspace{-5mm}
    \hfill\caption{
       \textbf{SpatialStack: Layered Geometry-Language Fusion.}
    Conventional VLMs (a) fuse only a single deep geometry feature with vision tokens, which limits both fine-grained spatial understanding and high-level spatial reasoning. SpatialStack (b) instead stacks multi-level geometry features and injects them hierarchically into successive LLM decoder layers, yielding stronger 3D spatial understanding across benchmarks.
    }
    \label{fig:teaser}
    \hfill \vspace{0mm}
}]

{
\renewcommand{\thefootnote}{\fnsymbol{footnote}}
\footnotetext[1]{Equal contribution.}
}

\begin{abstract}
Large vision-language models (VLMs) still struggle with reliable 3D spatial reasoning, a core capability for embodied and physical AI systems. This limitation arises from their inability to capture fine-grained 3D geometry and spatial relationships. While recent efforts have introduced multi-view geometry transformers into VLMs, they typically fuse only the deep-layer features from vision and geometry encoders, discarding rich hierarchical signals and creating a fundamental bottleneck for spatial understanding. To overcome this, we propose SpatialStack, a general hierarchical fusion framework that progressively aligns vision, geometry, and language representations across the model hierarchy. Moving beyond conventional late-stage vision-geometry fusion, SpatialStack stacks and synchronizes multi-level geometric features with the language backbone, enabling the model to capture both local geometric precision and global contextual semantics. Building upon this framework, we develop VLM-SpatialStack, a model that achieves state-of-the-art performance on multiple 3D spatial reasoning benchmarks. Extensive experiments and ablations demonstrate that our multi-level fusion strategy consistently enhances 3D understanding and generalizes robustly across diverse spatial reasoning tasks, establishing SpatialStack as an effective and extensible design paradigm for vision-language-geometry integration in next-generation multimodal physical AI systems.
\end{abstract}
\section{Introduction}
\label{sec:intro}
Understanding and reasoning about physical space are fundamental capabilities for any intelligent system that aims to perceive, communicate, and act in the physical world. Motivated by this, recent work on spatial reasoning aims to enable embodied agents to interpret scene layouts, predict interactions, and plan actions in 3D environments, forming a cognitive bridge between perception and action~\cite{fan2026vlm3r, wu2025spatial, zheng2025learning, liu2025spatial, yang2025visual, yang2025cambrian}. Despite remarkable progress in large vision-language models (VLMs), reliable spatial reasoning remains challenging, as these models often fail to effectively encode 3D geometry and spatial relationships and to associate them with language instructions, which are essential for everyday spatial tasks that require both low-level and high-level reasoning. For instance, they struggle to estimate relative distances in static scenes~\cite{yang2025thinking, li2025sti} and cannot reliably distinguish ``left'' from ``right'' when reasoning about motion in dynamic environments~\cite{zhou2025vlm4d}. In embodied AI applications such as robotic navigation, manipulation, and spatial assistance under XR, such limitations prevent VLMs from grounding their understanding in the complex and dynamic physical world.

Noticing these limitations of conventional VLMs, many recent works still prioritize image-level semantic alignment over the understanding of spatial and geometric structures~\cite{tang2025tulip, qi2025beyond, kamath2023what}. Bridging this gap requires unifying geometric awareness with vision-language reasoning within a single framework, which is a key step toward reliable spatial intelligence. This naturally raises a fundamental question: \textit{How can vision--language--geometry be effectively unified in VLMs to enable reliable spatial reasoning?}
An initial line of work sought to compensate for these weaknesses by integrating explicit geometric inputs (e.g., pre-computed point clouds or depth maps) into VLMs. For instance, early models like 3D-LLM~\cite{hong20233d} and LEO~\cite{huang2023embodied} used external point cloud encoders, while later methods like LLaVA-3D~\cite{zhu2025llava} and Video-3D LLM~\cite{zheng2025video} introduced lightweight encoders for RGB-D fusion. However, the reliance on these external, pre-processed inputs significantly limits their applicability. In parallel, rapid advancements in end-to-end multi-view geometry transformers, including DUST3R~\cite{wang2024dust3r}, CUT3R~\cite{wang2025continuous}, and VGGT~\cite{wang2025vggt}, have provided a more unified and powerful alternative to map uncalibrated images to 3D point maps. These models can infer rich geometric attributes such as depth, camera pose, and 3D structure directly from multi-view images, thereby bypassing traditional, computationally expensive geometric pipelines (e.g., Structure-from-Motion~\cite{schonberger2016structure}). Inspired by this progress, recent multimodal models, such as Spatial-MLLM~\cite{wu2025spatial}, VLM-3R~\cite{fan2026vlm3r}, and VG-LLM~\cite{zheng2025learning}, have begun integrating these geometry encoders into VLM frameworks, showing initial promise in improving spatial reasoning.

Nevertheless, most of these integrations focus only on fusing the final-layer features of geometry transformers with features from vision encoders. This is a critical limitation, as many geometry encoders adopt the DPT architecture~\cite{ranftl2021vision}, which explicitly extracts multi-level representations from different transformer layers to recover detailed geometric information. At the same time, a generalizable spatial-visual fusion mechanism has to account for hierarchical real-world tasks, ranging from low-level depth estimation and surface reconstruction to high-level relational reasoning and goal-directed planning. By sampling only the last layer, existing models discard the rich hierarchical geometric cues embedded in intermediate layers and overlook how different levels of geometric and semantic features contribute to spatial reasoning. Unsurprisingly, this single-level fusion design can improve performance on specific spatial benchmarks but creates a bottleneck that fundamentally constrains 3D understanding.

In this paper, we are motivated by the hierarchical nature of spatial reasoning tasks in 3D environments, and we systematically study how fusion layers across vision encoders, geometry encoders, and large language model (LLM) decoders affect multimodal spatial reasoning. Our analysis first shows that geometry-language fusion in multimodal LLMs follows a hierarchical pattern similar to vision encoding: shallow features enhance fine-grained spatial perception, while deeper features support high-level contextual reasoning. Building on these insights, we introduce SpatialStack, a general hierarchical fusion framework that integrates multi-level geometric features into multimodal LLMs. As shown in~\cref{fig:teaser}, unlike prior methods that fuse geometry only at deep encoder layers, SpatialStack progressively aligns geometric and language representations throughout the model hierarchy, capturing both detailed local geometry and global semantic context. Extensive experiments on multiple benchmarks demonstrate that our approach significantly improves 3D spatial reasoning, achieving strong performance on tasks requiring both detailed perception and holistic spatial understanding.

We summarize our \textbf{contributions} as follows:

\begin{itemize}
\item We present the first systematic analysis of how fusion layers across vision encoders, geometry encoders, and LLM decoders affect the granularity of spatial reasoning. Our layer-wise study reveals a hierarchical geometry--language correspondence, where shallow layers capture fine spatial details and deeper layers encode global structure and context.
\item We propose \textbf{SpatialStack}, a general hierarchical fusion framework that progressively aligns multi-level geometric and language features. This design goes beyond conventional final-stage vision-language fusion and supports joint reasoning over local and global spatial context.
\item While SpatialStack is model-agnostic and can be applied to any base multimodal LLM, we develop \textbf{VLM-SpatialStack} as a concrete realization using the Qwen series. Extensive experiments and ablation studies across multiple benchmarks show that SpatialStack achieves state-of-the-art performance and strong generalization on diverse 3D spatial reasoning tasks.
\end{itemize}
\section{Related Work}
\label{sec:related_work}

\begin{figure*}[t]
  \centering
   \includegraphics[width=0.8\linewidth]{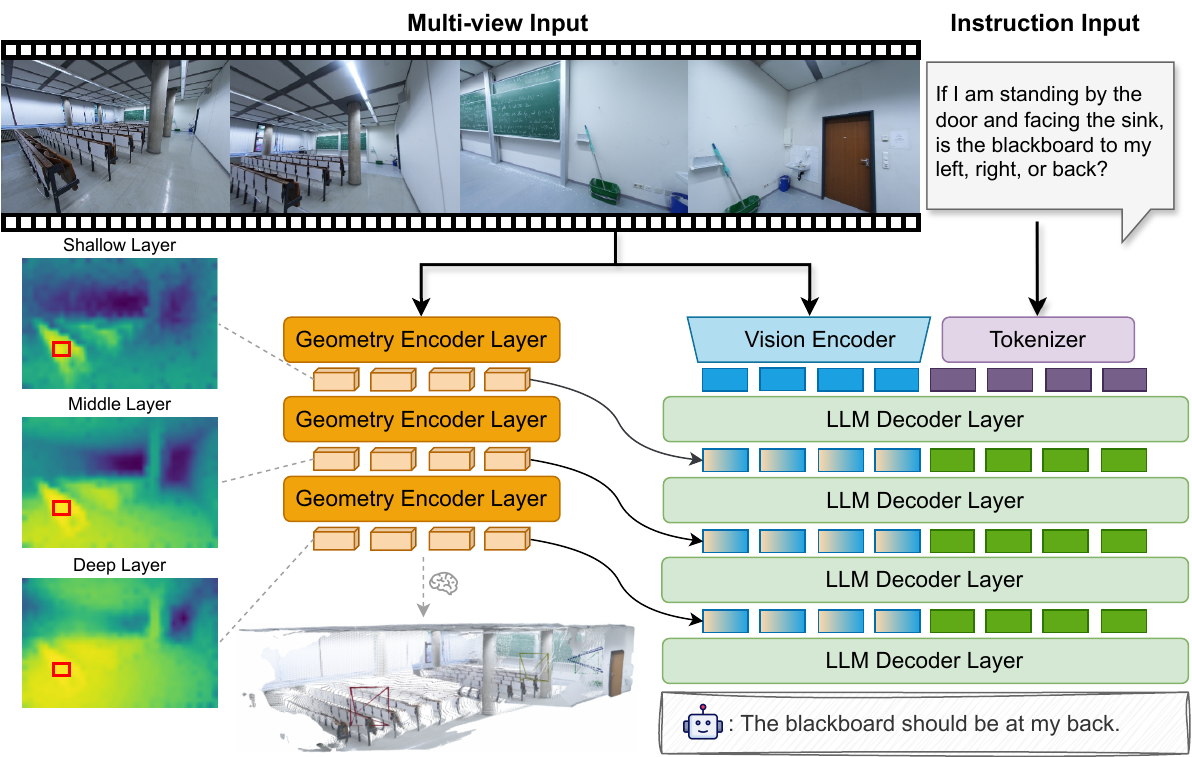}
    \caption{
\textbf{Architecture of SpatialStack.}
A standard VLM backbone is coupled with a multi-view geometry encoder whose layer-wise features are processed by layer-specific projectors and sequentially injected into the LLM decoder, progressively integrating geometric cues. Explanation of the similarity heatmaps on the left is provided in Sec.~\ref{sec:why_multilevel_geo}. This multi-level injection preserves both fine-grained geometric structure and high-level spatial context, supporting more reliable low-level understanding and high-level reasoning.}
   \label{fig:pipeline}
   \vspace{-1.2mm}
\end{figure*}

\paragraph{Large Multimodal Models (MLLMs)} Early works such as CLIP~\cite{radford2021learning} demonstrated the efficacy of learning joint vision-language representations from web-scale image-text pairs through contrastive pre-training. This paradigm was extended by subsequent models like Flamingo~\cite{alayrac2022flamingo}, which bridged powerful pre-trained vision and language models, and the BLIP series~\cite{li2022blip, li2023blip2}, which introduced bootstrapping methods and lightweight querying transformers to unify understanding and generation. A significant advancement in MLLM development was the advent of visual instruction tuning, effectively employed by models such as InstructBLIP~\cite{dai2023instructblip} to enhance instruction-following capabilities. Models like Qwen2.5-VL~\cite{bai2025qwen2.5vl}, LLaVA~\cite{liu2023visual} and MiniGPT-4~\cite{zhu2023minigpt} popularized a simple and effective architecture for this tuning, connecting a pre-trained vision encoder to a large language model (LLM) using only a simple projection layer. This simple design has spurred research into more effective fusion strategies, such as exploring different visual encoders~\cite{jiang2023clip} or deeper token stacking~\cite{meng2024deepstack}. This architectural blueprint, while powerful for general-purpose multimodal chat and semantic understanding, established a paradigm of fusing only the final-layer visual features with the language backbone. Consequently, as noted by recent analyses~\cite{kamath2023what, qi2025beyond, tang2025tulip}, these models are trained primarily for semantic alignment and often fail to capture the fine-grained spatial and geometric structures essential for physical reasoning.

\vspace{-4mm}
\paragraph{Spatial Reasoning in Vision-Language Models} The limitations of standard MLLMs in spatial reasoning have been well-documented, prompting recent efforts to quantify these deficiencies through benchmarks like VSI Bench~\cite{yang2025thinking}, Spar Bench~\cite{zhang2025flatland}, BLINK~\cite{fu2024blink}, and Cambrian-1~\cite{tong2024cambrian}. Cambrian-S~\cite{yang2025cambrian} further demonstrated a lack of ``implicit 3D spatial cognition", while VLM4D~\cite{zhou2025vlm4d} was the first to highlight the challenges of spatiotemporal (4D) reasoning in dynamic scenarios, followed by more recent efforts on dynamic 4D understanding and world modeling~\cite{huang2026thinkingdynamicsmultimodallarge,wen2025dynamicverse}. To address these gaps, one line of work focused on injecting explicit 3D data, such as 3D-LLM~\cite{hong20233d} which processes point clouds, or more lightweight approaches like LLaVA-3D~\cite{zhu2025llava} and Video-3D LLM~\cite{zheng2025video} that endow MLLMs with 3D awareness, with applications in embodied tasks~\cite{huang2023embodied}. A different strategy enhanced spatial abilities through novel training paradigms. Spatial-SSRL~\cite{liu2025spatial} introduced a self-supervised reinforcement learning framework, and Visual Spatial Tuning (VST)~\cite{yang2025visual} proposed a comprehensive tuning framework with a large-scale dataset (VST-P) and a progressive pipeline. These latter methods enhance spatial intelligence but primarily focus on training objectives and data augmentation rather than the core fusion architecture.

\vspace{-4mm}
\paragraph{Vision-Language-Geometry Fusion} The integration of explicit geometric reasoning within MLLMs has been recently catalyzed by the advent of powerful, feed-forward geometry encoders. Models such as DUST3R series~\cite{wang2024dust3r, leroy2024grounding} and CUT3R~\cite{wang2025continuous} can infer dense, consistent point maps from unposed multi-view images, while VGGT~\cite{wang2025vggt} introduced a unified transformer to predict diverse 3D attributes from video. The availability of these rich geometric features has inspired two parallel fusion paradigms. One line of work focuses on building \textit{explicit} spatial semantic representations, such as methods that distill 2D image or video foundation model features into 3D or 4D explicit feature field representations~\cite{zhou2024feature, thai2025splattalk, zhou2025feature4x}, build queryable 3D world models by fusing pixel-aligned features into 3D maps~\cite{jatavallabhula2023conceptfusion}, or map images directly to semantic radiance fields~\cite{fan2024large}. A parallel approach, which our work follows, \textit{implicitly} fuses geometric priors into the latent space of the MLLM. Recent models have shown initial promise in this direction: Spatial-MLLM~\cite{wu2025spatial} employs a dual-encoder architecture, VG-LLM~\cite{zheng2025learning} fuses features at the patch level, VLM-3R~\cite{fan2026vlm3r} introduces a cross-attention mechanism, and SSR~\cite{liu2025ssr} focuses on rationale-guided fusion. However, as identified in our analysis, these integrations typically fuse only the final-layer features from the geometry and vision encoders~\cite{wu2025spatial, zheng2025learning}. This single-level fusion design discards the rich, hierarchical geometric cues embedded in intermediate layers, creating a fundamental bottleneck for fine-grained spatial reasoning. Our work, SpatialStack, directly addresses this limitation by introducing a hierarchical fusion framework that progressively aligns multi-level geometry features with the language backbone.

\section{How Multi-level Geometry Features Facilitate Spatial Reasoning}
\label{sec:why_multilevel_geo}

\paragraph{Qualitative Analysis}
To validate our motivation, we begin by examining why relying solely on deep-layer geometry features is insufficient. As illustrated in~\cref{fig:pipeline}, we take one input view and unflatten the tokens from different layers of the geometry encoder back into their original $H \times W$ spatial layout. We then select a small patch (red bounding box) as the region of interest (ROI) and compute patch-wise similarity maps between the ROI and all other spatial locations: brighter regions indicate higher similarity, and darker regions indicate lower similarity. We observe a clear trend: shallow layers retain sharp local structures and well-defined geometric boundaries, while deeper layers produce overly homogeneous activations, where many regions appear similar in latent space despite having distinct physical geometry. This mismatch suggests that deep geometry features lose fine-grained spatial cues critical for reasoning about scene layout and spatial relations. These findings motivate our approach: leveraging multi-level geometry features, especially shallow-layer cues, to enrich spatial grounding and improve fine-grained 3D spatial reasoning in VLMs.

\begin{figure}[t!]
    \centering
    \includegraphics[width=0.95\linewidth]{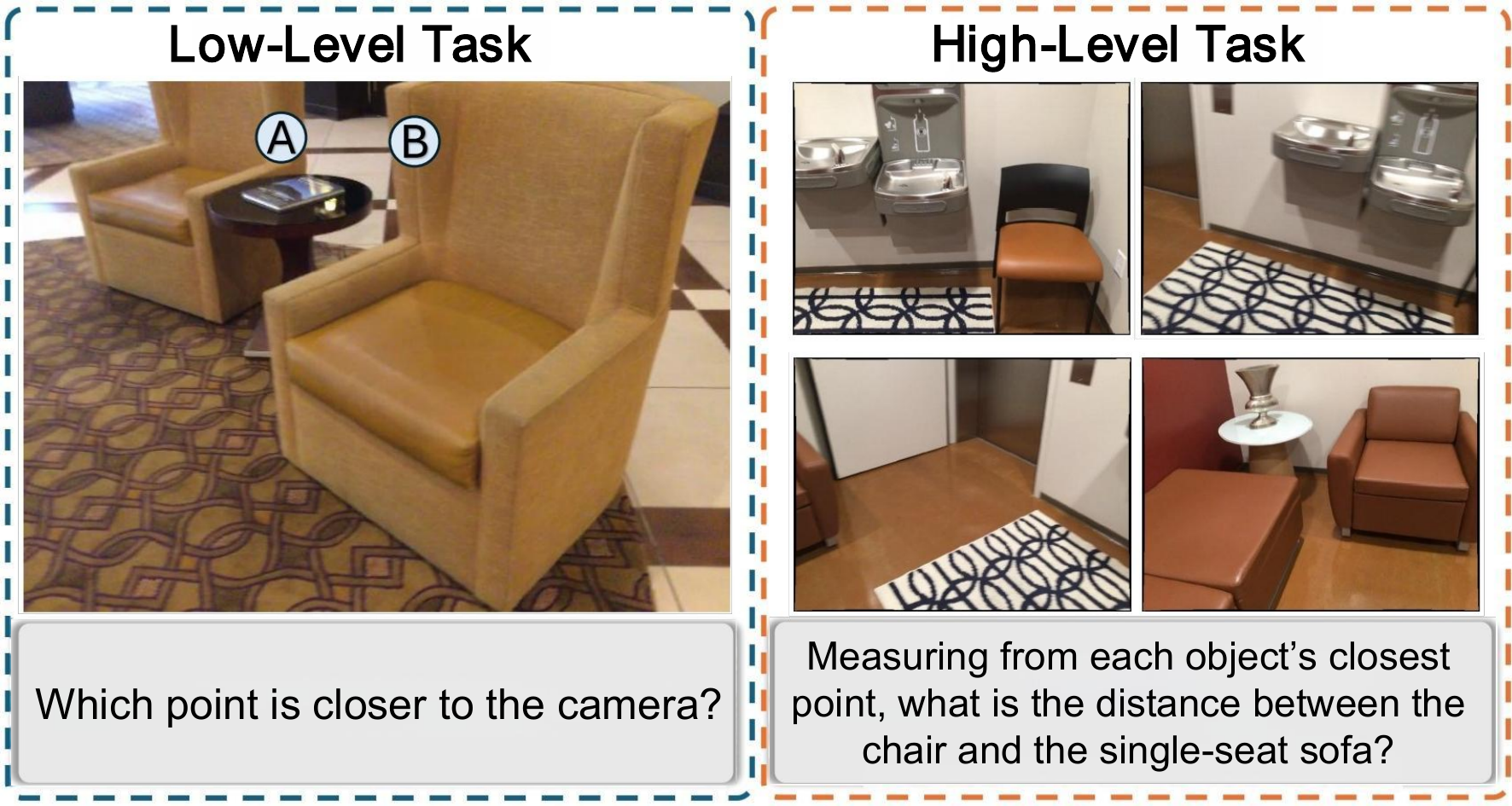}
    \caption{\textbf{Examples of spatial tasks at different levels.}
    The left example (\textit{Low-Level Task}) targets fine-grained geometric perception, such as determining which of two points is closer to the camera. 
    The right example (\textit{High-Level Task}) requires higher-level spatial reasoning, where the model must estimate the distance between two objects by comparing their closest points in 3D space.
    }
    \vspace{-2mm}
    \label{fig:low_high_level_tasks}
\end{figure}

\paragraph{Quantitative Analysis}
We further investigate how geometric features from different layers influence spatial reasoning performance. Firstly, we follow the difficulty hierarchy defined in SPAR~\cite{zhang2025flatland} dataset, which categorizes spatial tasks based on the required complexity of spatial understanding. SPAR divides spatial tasks into three cognitive levels: \textit{perception (low)}, \textit{reasoning (medium)}, and \textit{imagination (high)}. Low-level tasks emphasize fundamental geometric perception, such as single-view depth estimation and distance comparison based on local pixel/feature cues; high-level tasks require aggregating spatial information across multiple viewpoints for global spatial reasoning, such as cross-view object spatial relations and path reasoning (see \cref{fig:low_high_level_tasks}). 

Based on this criterion, our \textbf{low-level tasks} evaluating fundamental perception include BLINK's \textit{relative depth}~\cite{fu2024blink} and specific tasks in SPAR-Bench~\cite{zhang2025flatland}: depth (\textit{Depth-OC/OO/OC-MV/OO-MV}) and absolute distance (\textit{Dist-OC/OO/OC-MV/OO-MV}) (see~\cite{zhang2025flatland} for more details). Conversely, all \textit{VSI-Bench} tasks are categorized as \textbf{high-level tasks}, as they require complex multi-view spatial fusion and 3D relational reasoning. We do not include SPAR’s medium or high tasks in this study, as we aim to establish a clearer two-level comparison that isolates the distinct effects of geometric feature integration on \textit{basic perception ability} versus \textit{complex spatial reasoning ability}.

Under the task definitions above, we further conduct a quantitative analysis of the performance impact of injecting geometric features from different layers into VLMs. Specifically, following VG-LLM~\cite{zheng2025learning}, we extract geometric features from a single layer of the geometry encoder (VGGT~\cite{wang2025vggt}), project them through a projector, and add them to the last-layer features of the vision encoder. We denote this geometry-vision fusion as GVF in the rest of our paper. 
The fused geometry-vision features are then concatenated with text tokens and fed into the LLM decoder. We experiment with injecting geometric features from the 4th, 11th, 17th, and 23rd layers, and evaluate the performance on the two task levels.

\begin{figure}[t!]
    \centering
    \includegraphics[width=1.0\linewidth]{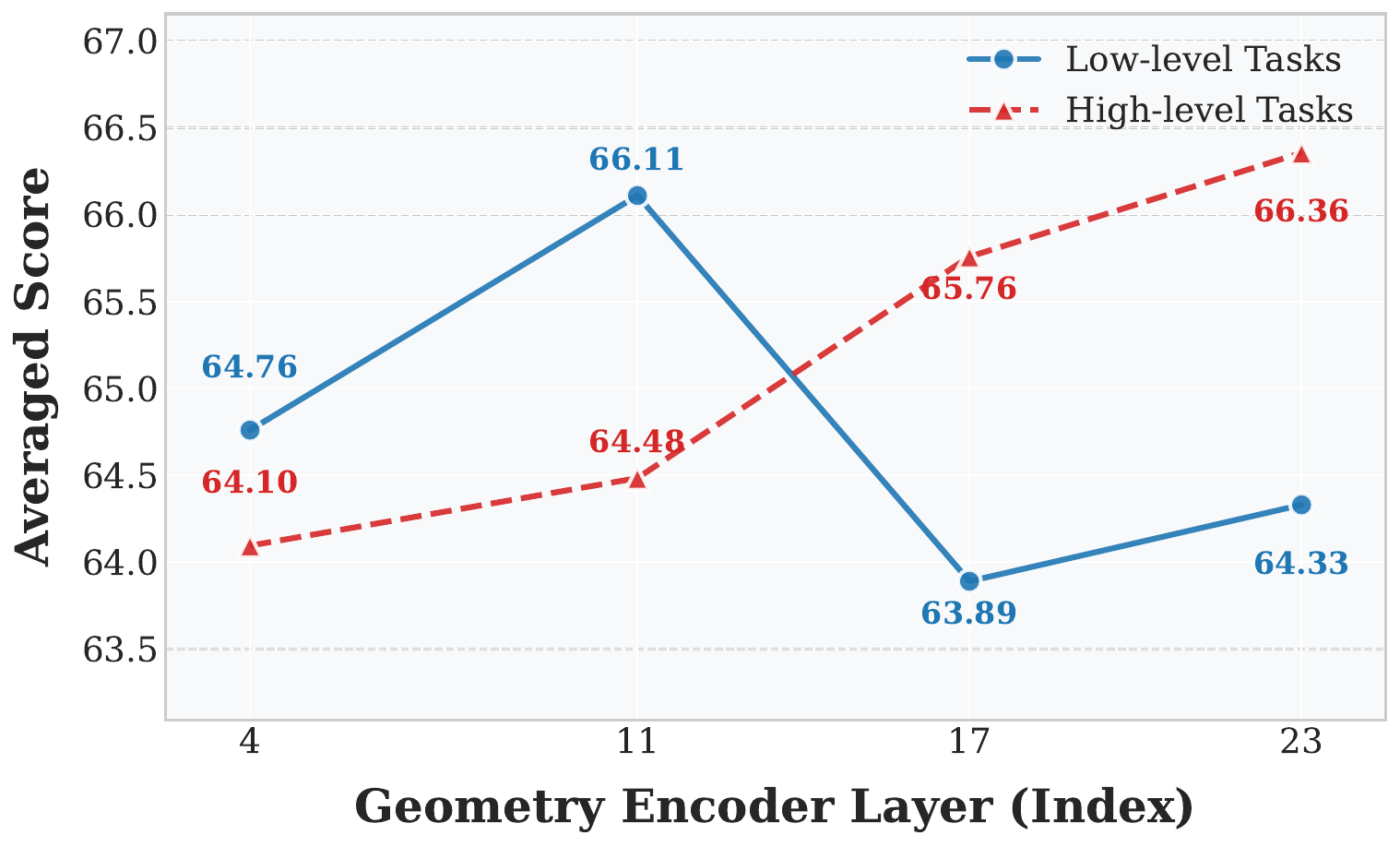}
    \caption{\textbf{Effect of Geometry Injection Layers on Spatial Tasks.}
    Deeper layers improve high-level tasks, while low-level tasks peak at layer 11 and decline at deeper layers, suggesting a trade-off between fine-grained perception and higher-level reasoning.
    }
    \vspace{-1mm}
    \label{fig:geo_layer_effect}
\end{figure}

As shown in~\cref{fig:geo_layer_effect}, the results demonstrate that the choice of injection layer has a significant impact on different levels of spatial tasks: as the injection layer becomes deeper, the performance on low-level tasks declines, while the performance on high-level tasks improves significantly. This phenomenon suggests that geometric features from different layers play distinct roles in spatial understanding: features from shallower layers provide fine-grained local geometric cues beneficial for basic spatial perception, whereas deeper features encode more global structural and semantic relationships, making them more suitable for complex spatial reasoning.

Given the complementary strengths of shallow and deep features, a multi-layer fusion strategy should intuitively enhance both perception and reasoning. To test this, \cref{tab:geo_depth_avg} compares various fusion strategies against Qwen3.5~\cite{qwen35blog}, the base model fine-tuned on our spatial reasoning datasets without any geometric enhancements. Surprisingly, naive multi-layer fusion fails to achieve the best. Instead, it yields a compromised performance, falling behind the 11th-layer single-fusion on low-level tasks and the 23rd-layer on high-level tasks. This sub-optimal outcome reveals that simply adding hierarchical features into the vision pathway causes feature interference rather than synergy. This highlights that merely extracting multi-level cues is insufficient; the true challenge lies in the fusion strategy---a realization that serves as the primary catalyst for SpatialStack.

\begin{table}[t]
\centering
\footnotesize
\setlength{\tabcolsep}{4pt}
\renewcommand{\arraystretch}{1.05}
\resizebox{\linewidth}{!}{
\begin{tabular}{l|c|c|c}
\toprule
\textbf{Model} & \textbf{Low-Level Avg} & \textbf{High-Level Avg} & \textbf{Overall} \\
\midrule
  Qwen3.5 (fine-tuned)         & 61.37 & 64.76 & 63.07 \\
  Single-layer (geo enc: 11)   & \textbf{66.11} & 64.48 & 65.30 \\
  Single-layer (geo enc: 17)   & 63.89 & 65.76 & 64.83 \\
  Single-layer (geo enc: 23)   & 64.33 & \textbf{66.36} & \textbf{65.35} \\
  Multi-Layer Fusion           & 64.69 & 65.15 & 64.92 \\
\bottomrule
\end{tabular}
}
\caption{\textbf{Ablation Results on Geometry Token Fusion Depth.} 
Simply fusing multi-layer geometry features to the visual features yields suboptimal performance, while selecting an appropriate single geometry encoder layer achieves better task-specific trade-offs.
}
\label{tab:geo_depth_avg}
\end{table}
\section{Where to fuse Multi-level Geometry Features}
\label{sec:method}

The observation of feature interference during naive vision-pathway fusion in \cref{sec:why_multilevel_geo} prompts a critical question: \textit{where} and \textit{how} should these hierarchical geometric features be integrated to maximally enhance a VLM's spatial reasoning? Should they be confined to the vision encoder, or directly injected into the language model? 

\subsection{SpatialStack: Geometry-Language Fusion}
While most prior works~\cite{zheng2025learning, wu2025spatial, fan2026vlm3r} confine geometric enhancements to the vision encoder, we hypothesize that injecting these features directly into the Large Language Model (LLM) provides a more flexible, high-capacity space for multi-scale spatial reasoning. Inspired by DeepStack's~\cite{meng2024deepstack} success in stacking visual tokens within the LLM, we shift geometry integration to the language side and propose SpatialStack: a novel, first-of-its-kind layered geometry–language fusion framework.

As shown in~\cref{fig:pipeline}, SpatialStack performs multi-level fusion between a geometry encoder and an LLM decoder. Its key idea is to inject geometric features from multiple layers of the geometry encoder into corresponding layers of the LLM, forming a hierarchy of geometric representations. This progressive stacking introduces geometric cues throughout the decoding process, strengthening spatial grounding and improving reasoning across tasks of varying difficulty. 

Importantly, SpatialStack is a general framework that can be integrated with any open-source VLM. We instantiate VLM-SpatialStack using the latest Qwen3.5~\cite{qwen35blog} as our primary base model. To ensure a fair comparison with existing baselines~\cite{wu2025spatial, zheng2025learning, yang2025cambrian}, we also provide an instantiation based on the same base model they use, Qwen2.5-VL~\cite{bai2025qwen2.5vl}.

\subsection{VLM-SpatialStack}
\paragraph{VLM Architecture.} 
Given $K$ input frames $\{\mathbf{I}_k \in \mathbb{R}^{H\times W\times3}\}_{k=1}^{K}$,  
each frame is encoded by a shared vision encoder into tokens  
$\mathbf{V}_k \in \mathbb{R}^{N \times D_{\text{vis}}}$,  
where $p$ is the patch size and $N=\tfrac{H}{p}\!\times\!\tfrac{W}{p}$ is the number of patch tokens per frame.  
A spatial merger groups every $2\times2$ neighboring patches (stride factor $s=2$),  
producing $\tilde{\mathbf{V}}_k \in \mathbb{R}^{N' \times D_{\text{lang}}}$  
with $N'=\tfrac{N}{s^2}=\tfrac{HW}{(ps)^2}$.  
Merged tokens from all frames are concatenated along the sequence dimension:
\[
\tilde{\mathbf{V}} = [\,\tilde{\mathbf{V}}_1;\ldots;\tilde{\mathbf{V}}_K\,] 
\in \mathbb{R}^{(K N') \times D_{\text{lang}}}.
\]
The concatenated visual tokens $\tilde{\mathbf{V}}$ and text tokens  
$\mathbf{T} \in \mathbb{R}^{M \times D_{\text{lang}}}$  
form the multimodal input sequence 
$\mathbf{H}_0 = [\,\tilde{\mathbf{V}};\mathbf{T}\,]$.  
This sequence is then processed by $L$ stacked transformer layers in the LLM decoder:
\begin{equation}
\mathbf{H}_L^{\text{llm}} 
= f_L^{\text{llm}}\!\Big(
    f_{L-1}^{\text{llm}}\!\big(
        \cdots f_1^{\text{llm}}(\mathbf{H}_0)
    \big)
\Big),
\label{eq:llm_forward}
\end{equation}
where $\mathbf{H}_L^{\text{llm}}$ denotes the final hidden representations produced by the LLM decoder for downstream prediction.

\paragraph{Geometry Encoder.}
We employ the Visual Geometry Grounded Transformer (VGGT)~\cite{wang2025vggt} as our geometry encoder. 
Given the same set of $K$ input images 
$\{\mathbf{I}_k \in \mathbb{R}^{H \times W \times 3}\}_{k=1}^{K}$, 
each image is divided into non-overlapping patches of size $p\times p$, 
resulting in $N=(H/p)\!\times\!(W/p)$ patch tokens. 
In addition to the patch tokens, VGGT includes camera and register tokens to encode view-specific and shared geometric context. 
The initial token sequence for view $k$ is thus
\begin{equation}
\mathbf{Z}_0^{(k)} = [\,\mathbf{c}_k;\,\mathbf{r}_k;\,\mathbf{p}_k\,] 
\in \mathbb{R}^{(1+R+N)\times D_{\text{geo}}},
\label{eq:vggt_tokens}
\end{equation}
where $\mathbf{p}_k$ denotes the patch tokens of image $\mathbf{I}_k$. 
All view-specific sequences are concatenated and jointly processed by $L$ stacked transformer layers:
\begin{equation}
\mathbf{Z}_L = f_L^{\text{geo}}\!\Big(
    f_{L-1}^{\text{geo}}\!\big(
        \cdots f_1^{\text{geo}}([\mathbf{Z}_0^{(1)};\ldots;\mathbf{Z}_0^{(K)}])
    \big)
\Big),
\label{eq:vggt_forward}
\end{equation}
where $f_l^{\text{geo}}(\cdot)$ denotes the $l$-th transformer layer in VGGT.
While the original VGGT employs Dense Prediction Transformer (DPT) heads~\cite{ranftl2021vision} 
for outputs such as depth, point clouds, and camera parameters, 
we instead extract intermediate hidden states $\mathbf{Z}_l$ from selected layers 
as multi-view geometric features for fusion with the vision--language model.

\paragraph{Layered Geometry--Language Fusion.} As illustrated in ~\cref{fig:pipeline}, 
we extract multi-level patch features $\mathbf{Z}_{l_i}$ 
from the geometry encoder defined in~\cref{eq:vggt_forward}.  
Specifically, we take the patch-token outputs of the $l_i$-th layers 
($l_i \in \{11, 17, 23\}$, counted from zero) of VGGT after removing camera and register tokens, 
thereby yielding $\mathbf{Z}_{l_i} \in \mathbb{R}^{(K N)\times D_{\text{geo}}}$ 
that represent geometric information at different, progressively richer abstraction levels.
Each feature $\mathbf{Z}_{l_i}$ is processed by a layer-specific geometry token merger
$\mathcal{M}_{\text{geo}}^{(l_i)}$
to align its spatial resolution and embedding dimension with that of $\mathbf{H}$:
\begin{equation}
\mathbf{G}_{l_i} = \mathcal{M}_{\text{geo}}^{(l_i)}(\mathbf{Z}_{l_i}),
\quad
\mathbf{G}_{l_i} \in \mathbb{R}^{N' \times D_{\text{lang}}}.
\label{eq:geo_merger}
\end{equation}

Finally, the geometry features
$\{\mathbf{G}_{l_1}, \mathbf{G}_{l_2}, \mathbf{G}_{l_3}\}$
extracted from VGGT layers $\{11, 17, 23\}$ 
are injected into LLM decoder layers 
$\{0, 1, 2\}$ as additive residuals:
\begin{equation}
\mathbf{H}^{(j)'} = \mathbf{H}^{(j)} + \mathbf{G}_{l_j},
\quad
j \in \{0,1,2\}.
\label{eq:geo_inject}
\end{equation}

\paragraph{Optimization.}
We train the entire model under a single objective, the next-token negative log-likelihood (cross-entropy):
\begin{equation}
\mathcal{L}_{\text{ce}}(\theta)
= - \sum_{i=1}^{|o|} \log P_\theta\!\big( o^{(i)} \,\big|\, o^{(<i)},\, q,\, \mathcal{C} \big),
\label{eq:inst_ce}
\end{equation}
where $q$ denotes the system prompt and question, $o^{(i)}$ is the $i$-th token of the ground-truth answer, $o^{(<i)}$ are the preceding answer tokens, and $\mathcal{C}$ represents the multimodal context (e.g., input frames). During instruction tuning, we freeze both the vision encoder and the geometry encoder, and update only the multimodal fusion modules and the LLM decoder. This choice preserves the pretrained visual and geometric representations while allowing the model to learn how to align and integrate them effectively for spatial reasoning. No auxiliary objectives or task-specific losses are introduced; spatial priors emerge purely through unified instruction tuning across diverse spatial tasks.

\subsection{SpatialStack vs. Geometry-Vision Fusion}

To evaluate the effectiveness of SpatialStack, we compare it against three baselines: base model Qwen3.5, a naive single-layer Geometry–Vision Fusion (GVF-L23) equivalent to VG-LLM~\cite{zheng2025learning} built on Qwen3.5, and a naive multi-layer fusion (GVF-L11/17/23). Across four spatial reasoning benchmarks in~\cref{tab:ablation_cross_bench}, SpatialStack achieves the best overall average and obtains the highest scores on \textit{VSI-Bench}~\cite{yang2025thinking}, \textit{SPAR-Bench}~\cite{zhang2025flatland}, and \textit{CV-Bench}~\cite{tong2024cambrian}. While the base Qwen3.5 model remains strongest on \textit{BLINK-Spatial}~\cite{fu2024blink}, the naive geometry-vision fusion approaches (GVF-L23 and GVF-L11/17/23) suffer severe performance drops on this dataset and fail to generalize effectively across tasks. These results highlight that straightforward visual-pathway injection lacks robust generalization, whereas SpatialStack demonstrates superior cross-task transfer ability.

\begin{table}[t]
  \centering
  \resizebox{0.85\linewidth}{!}{
  \begin{tabular}{lccccc}
  \toprule
  Methods &
  \rotatebox{60}{VSI-Bench} &
  \rotatebox{60}{SPAR-Bench} &
  \rotatebox{60}{BLINK-Spatial} &
  \rotatebox{60}{CV-Bench} &
  \rotatebox{60}{Overall} \\
  \midrule
  Qwen3.5 (fine-tuned)   & 64.76 & 68.75 & \cellcolor{gray!20}\textbf{56.10} & 84.49 & 68.52 \\
  GVF-L23 (VG-LLM~\cite{zheng2025learning})      & 66.36 & 70.83 & 51.91 & 84.64 & 68.43 \\
  GVF-L11/17/23 & 65.15 & 71.20 & 51.28 & 84.33 & 67.99 \\
  SpatialStack & \cellcolor{gray!20}\textbf{67.52} & \cellcolor{gray!20}\textbf{71.39} & 52.12 & \cellcolor{gray!20}\textbf{85.53} &
  \cellcolor{gray!20}\textbf{69.14} \\
  \bottomrule
  \end{tabular}}
  \vspace{-3mm}
  \caption{\textbf{Cross-benchmark Ablation.}
  SpatialStack achieves the best cross-task transfer ability, obtaining the highest scores on \textbf{VSI-Bench}, \textbf{SPAR-Bench},
  \textbf{CV-Bench}, and the overall average, while the \texttt{Qwen3.5} baseline remains strongest on \textbf{BLINK-Spatial}.
  \cellcolor{gray!20} Gray cells denote the highest value in each column.
  }
  \label{tab:ablation_cross_bench}
  \vspace{-3mm}
  \end{table}
\section{Experiments}

\begin{figure*}[ht!]
    \captionsetup{type=table}
    \vspace{-0.2cm}
    \centering
    \begin{minipage}{0.64\textwidth}
    \centering
    \fontsize{4.6pt}{4.4pt}\selectfont
    \setlength\tabcolsep{3pt}
    \renewcommand{\arraystretch}{1.2}
    \scalebox{1.4}{
    \begin{tabular}{r|cc|cccccccc}
    & & &
    \rotatebox{75}{Obj. Count} &
    \rotatebox{75}{Abs. Dist.} &
    \rotatebox{75}{Obj. Size} &
    \rotatebox{75}{Room Size} &
    \rotatebox{75}{Rel. Dist.} &
    \rotatebox{75}{Rel. Dir.} &
    \rotatebox{75}{Route Plan} &
    \rotatebox{75}{Appr. Order} \\
    Methods & Rank & Avg. & \multicolumn{4}{c}{\cellcolor{orange!10}Numerical Answer} & \multicolumn{4}{c}{\cellcolor{yellow!10}Multiple-Choice Answer} \\
    \hline
    \rowcolor{navyblue!5}
    \multicolumn{1}{l|}{\textit{Baseline}} & & & & & & & & & & \\
    Chance Level (Random) & - & - & - & - & - & - & 25.0 & 36.1 & 28.3 & 25.0 \\
    Chance Level (Frequency) & - & 34.0 & 62.1 & 32.0 & 29.9 & 33.1 & 25.1 & 47.9 & 28.4 & 25.2 \\
    \hline
    \rowcolor{navyblue!5}
    \multicolumn{1}{l|}{\textit{Proprietary Models (API)}} & & & & & & & & & & \\
    GPT-4o & \cellcolor{violet!30}{2} & 34.0 & 46.2 & 5.3 & 43.8 & 38.2 & 37.0 & 41.3 & 31.5 & 28.5 \\
    Gemini-2.5 Pro & \cellcolor{violet!45}{1} & 51.5 & 43.8 & 34.9 & 64.3 & 42.8 & 61.1 & 47.8 & 45.9 & 71.3 \\
    \hline
    \rowcolor{navyblue!5}
    \multicolumn{1}{l|}{\textit{Open-source Models}} & & & & & & & & & & \\
    LongVILA-8B & 15 & 21.6 & 29.1 & 9.1 & 16.7 & 0.0 & 29.6 & 30.7 & 32.5 & 25.5 \\
    Qwen2.5-VL-3B & 14 & 28.7 & 33.1 & 19.4 & 17.4 & 24.8 & 37.3 & 44.3 & 31.4 & 22.0 \\
    VILA-1.5-8B & 13 & 28.9 & 17.4 & 21.8 & 50.3 & 18.8 & 32.1 & 34.8 & 31.0 & 24.8 \\
    LongVA-7B & 12 & 29.2 & 38.0 & 16.6 & 38.9 & 22.2 & 33.1 & 43.3 & 25.4 & 15.7 \\
    VILA-1.5-40B & 11 & 31.2 & 22.4 & 24.8 & 48.7 & 22.7 & 40.5 & 25.7 & 31.5 & 32.9 \\
    LLaVA-OneVision-7B & 10 & 32.4 & 47.7 & 20.2 & 47.4 & 12.3 & 42.5 & 35.2 & 29.4 & 24.4 \\
    LLaVA-Video-7B & 9 & 35.6 & 48.5 & 14.0 & 47.8 & 24.2 & 43.5 & 42.4 & 34.0 & 30.6 \\
    LLaVA-OneVision-72B & 8 & 40.2 & 43.5 & 23.9 & 57.6 & 37.5 & 42.5 & 39.9 & 32.5 & 44.6 \\
    LLaVA-Video-72B & 7 & 40.9 & 48.9 & 22.8 & 57.4 & 35.3 & 42.4 & 36.7 & 35.0 & 48.6 \\
    Spatial-MLLM-4B & 6 & 47.0 & 65.3 & 34.8 & 63.1 & 45.1 & 41.3 & 46.2 & 33.5 & 46.3 \\
    VG-LLM-4B & 5 & 47.3 & 66.0 & 37.8 & 55.2 & 59.2 & 44.6 & 45.6 & 33.5 & 36.4 \\
    Qwen3.5-4B & 4 & 53.6 & 56.5 & 36.5 & 67.5 & 53.8 & 60.3 & 57.5 & 34.0 & 62.3 \\
    Cambrian-S-3B & \cellcolor{violet!15}{3} & 57.3 & \cellcolor{orange!15}{70.7} & 40.6 & \cellcolor{orange!15}{68.0} & 46.3 & \cellcolor{orange!15}{64.8} & 61.9 & 27.3 & 78.8 \\
    SpatialStack-4B (Qwen2.5) & \cellcolor{violet!30}{2} & 60.9 & 69.2 & \cellcolor{orange!15}{45.4} & 63.0 & \cellcolor{orange!15}{63.2} & 57.9 & \cellcolor{orange!15}{68.4} & \cellcolor{orange!15}{40.2} & \cellcolor{orange!15}{79.6} \\
    SpatialStack-5B (Qwen3.5) & \cellcolor{violet!45}{1} & 67.5 & \cellcolor{orange!30}{71.0} & \cellcolor{orange!30}{55.6} & \cellcolor{orange!30}{69.1} & \cellcolor{orange!30}{68.2} & \cellcolor{orange!30}{67.3} & \cellcolor{orange!30}{84.1} & \cellcolor{orange!30}{41.2} & \cellcolor{orange!30}{83.5} \\
    \end{tabular}
}
    \end{minipage}
    \hfill
    \begin{minipage}[c]{0.35\textwidth}
        \centering
        \includegraphics[width=\textwidth]{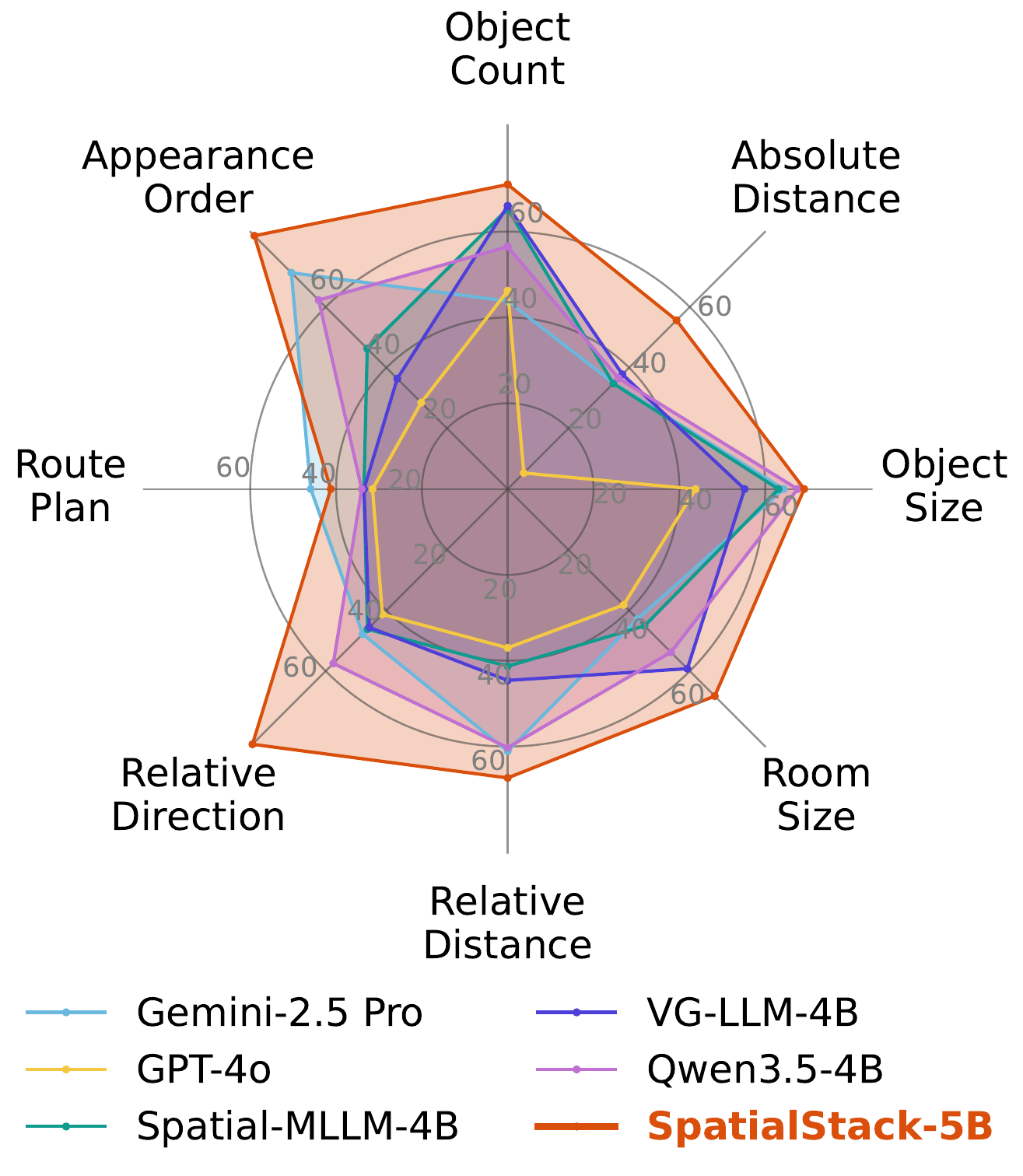}
    \end{minipage}
    \vspace{-0.2cm}
    \caption{\textbf{Evaluation on \bench.}
    \colorbox{orange!30}{Dark orange} cells denote the best \emph{open-source} result in each column,
    while \colorbox{orange!15}{light orange} cells denote the second-best \emph{open-source} result.
    Group-wise ranks within proprietary and open-source model blocks are highlighted in purple, with
    \colorbox{violet!45}{dark purple}, \colorbox{violet!30}{medium purple}, and \colorbox{violet!15}{light purple}
    indicating 1st, 2nd, and 3rd place, respectively.}
    \label{tab:main_table}
\end{figure*}

We describe our training setup in \cref{sec:implementation}, evaluate VLM-SpatialStack against state-of-the-art methods in \cref{sec:benchmark_comparison}, and provide extensive ablation studies in \cref{sec:ablation}.

\subsection{Training}
\label{sec:implementation}

\paragraph{Training Datasets Construction.}
Our training dataset is constructed by sampling from multiple spatial reasoning sources, including the SPAR and LLaVA-Hound subsets used in VG-LLM~\cite{zheng2025learning}, the ScanNet split adopted in VLM-3R~\cite{fan2026vlm3r}, and a selected portion of the VSI-590K corpus~\cite{yang2025cambrian}. SPAR~\cite{zhang2025flatland} provides large-scale spatial data generated from reconstructed scenes with 3D ground truth, while LLaVA-Hound~\cite{zhang2025direct} offers general-purpose video question–answering samples. VLM-3R reformulates spatial question–answer pairs in a VSI-Bench-style format, producing diverse reasoning tasks such as relative direction, object counting, and absolute distance estimation from real-world 3D–annotated scenes. We sample from these sources to ensure broad coverage of spatial reasoning types and maintain precise alignment between 3D geometry and textual descriptions.

\paragraph{Training Setup.}
We fine-tune the model using the standard language modeling cross-entropy loss. Training is performed with a batch size of 64 and a learning rate of $1\times10^{-5}$, optimized using the AdamW optimizer with a warmup ratio of 0.03 and a cosine learning rate schedule. During instruction tuning, the geometry encoder (VGGT) and the vision encoder are kept frozen, while the geometry token merger modules and the LLM decoder are trainable to learn geometry–language alignment.

\subsection{Evaluation}\label{sec:benchmark_comparison}

We evaluate our model on a diverse set of multimodal benchmarks that test both spatial and general reasoning, including VSI-Bench~\cite{yang2025thinking}, CV-Bench~\cite{tong2024cambrian}, SPAR-Bench~\cite{zhang2025flatland}, BLINK~\cite{fu2024blink}, and Video-MME~\cite{fu2025video}. These benchmarks cover a wide range of tasks, such as depth and distance estimation, object–relation reasoning, and video-based spatial understanding.

\paragraph{Evaluation on VSI-Bench.}
We evaluate our model on \textit{VSI-Bench}~\cite{yang2025thinking}, which contains over 5{,}000 QA pairs from egocentric indoor videos and includes both Multiple-Choice Answer (MCA) and Numerical Answer (NA) tasks. Following the official protocol, we report mean MCA accuracy and Mean Relative Accuracy for NA across confidence thresholds $C={0.5, 0.55, \ldots, 0.95}$. For comparison, we include representative proprietary models~\cite{hurst2024gpt, comanici2025gemini}, open-source video MLLMs~\cite{chen2024longvila, lin2024vila, zhang2024long, zhang2024video, li2024llava}, and geometry-aware methods at similar scales~\cite{yang2025cambrian, zheng2025learning, wu2025spatial}. 

As shown in~\cref{tab:main_table}, we demonstrate that SpatialStack serves as a general and highly effective paradigm for enhancing various VLMs. Applying our framework to both Qwen2.5~\cite{bai2025qwen2.5vl} and Qwen3.5~\cite{qwen35blog} yields substantial improvements over their untuned base models. Furthermore, under a fair comparison using the identical Qwen2.5 base model, SpatialStack significantly outperforms other concurrent geometry-aware MLLMs, such as Spatial-MLLM~\cite{wu2025spatial}, VG-LLM~\cite{zheng2025learning}, and Cambrian-S~\cite{yang2025cambrian}. Ultimately, our latest SpatialStack-5B (based on Qwen3.5) establishes a new state-of-the-art among all evaluated open-source models. Notably, despite lacking route-planning data during training, it still surpasses all open-source systems on this task, demonstrating robust zero-shot generalization for high-level spatial reasoning.

\vspace{-3mm}
\paragraph{Evaluation on CV-Bench.}
To assess 2D and 3D spatial perception, we evaluate on \textit{CV-Bench}~\cite{tong2024cambrian}, which measures performance via QA tasks constructed from standard vision datasets~\cite{zhou2017scene, lin2014microsoft, brazil2023omni3d}. We follow the official protocol and report average accuracy across all task types.
As shown in ~\cref{tab:cvbench_main}, our two versions of SpatialStack surpass all baselines of similar scale and same base models on both 2D and 3D subsets, demonstrating the benefits of multi-level geometry feature stacking for unified spatial perception.

\begin{table}[t]
\centering
\resizebox{1.0\linewidth}{!}{
\begin{tabular}{@{}r|cc|c@{}}
\toprule
\makecell[c]{\textbf{Model}} & \textbf{2D (\%)} & \textbf{3D (\%)} & \textbf{Avg. (\%)}\\
\midrule

\rowcolor{navyblue!5}
\multicolumn{4}{l}{\textit{Proprietary Models (API)}} \\
GPT-4o~\cite{hurst2024gpt} & 74.8 & 83.0 & 78.9 \\

\rowcolor{navyblue!5}
\multicolumn{4}{l}{\textit{Open-source Models}} \\
Mini-Gemini-HD-34B~\cite{li2024mini} & 71.5 & 79.2 & 75.4 \\
LLaVA-NeXT-34B~\cite{li2024llava} & 73.0 & 74.8 & 73.9 \\
Cambrian-1-34B~\cite{tong2024cambrian} & 74.0 & 79.7 & 76.9 \\
SAT-LLaVA-Video-7B~\cite{ray2024sat} & 73.0 & 83.8 & 78.4 \\
SPAR-8B~\cite{zhang2025flatland} & 72.3 & 89.1 & 80.7 \\
Qwen2.5-VL-3B~\cite{bai2025qwen2.5vl} & 67.9 & 70.4 & 69.2 \\
Qwen3.5-4B~\cite{qwen35blog} & \textbf{79.7} & 90.2 & 85.0 \\
Cambrian-S-3B~\cite{yang2025cambrian} & 76.1 & 76.3 & 76.2 \\

\rowcolor{navyblue!5}
\multicolumn{4}{l}{\textit{Dual-Encoder MLLMs}} \\
VG-LLM-4B~\cite{zheng2025learning} & 71.3 & 87.7 & 79.5 \\
SpatialStack-4B (Qwen2.5) & 75.4 & 87.0 & 81.2 \\
SpatialStack-5B (Qwen3.5) & 78.9 & \textbf{92.2} & \textbf{85.5} \\

\bottomrule
\end{tabular}
}
\vspace{-2mm}
\caption{
\textbf{Comparison on CV-Bench.}
Built on Qwen2.5, SpatialStack-4B outperforms its base model alongside VG-LLM and Cambrian-S. Scaling to Qwen3.5, SpatialStack-5B further improves upon its baseline to set a new state-of-the-art.
}
\label{tab:cvbench_main}
\end{table}

\paragraph{Evaluation on General-purpose Capabilities.}
We evaluate SpatialStack on a comprehensive suite of benchmarks: MMBench~\cite{liu2024mmbench} and Video-MME (general multi-modal/video understanding), BLINK (fine-grained visual perception), and TempCompass~\cite{liu2024tempcompass} (spatial-temporal reasoning). \cref{tab:general_eval} shows that our method maintains robust general capabilities while specializing in spatial-temporal tasks, confirming no catastrophic forgetting.

\begin{table}[t]
    \centering
    \resizebox{1.0\linewidth}{!}{ 
        \renewcommand{\arraystretch}{0.85}
        \setlength{\tabcolsep}{5pt}
        \begin{tabular}{l|cccc|c}
            \toprule
            Method & 
            \begin{tabular}[c]{@{}c@{}}MMBench\end{tabular} & 
            \begin{tabular}[c]{@{}c@{}}Video\\-MME\end{tabular} & 
            BLINK & 
            \begin{tabular}[c]{@{}c@{}}Temp\\Compass\end{tabular} & 
            Overall \\
            \midrule
            Qwen3.5-4B
            & 83.25
            & 62.44
            & \textbf{61.12}
            & 66.84
            & \textbf{68.41} \\
            SpatialStack-5B (Qwen3.5)
            & \textbf{83.42}
            & \textbf{63.74}
            & 55.46
            & \textbf{69.37}
            & 68.00 \\
            \bottomrule
        \end{tabular}
    }
    \vspace{-2mm} 
    \caption{\textbf{General Capabilities Evaluation.} Our SpatialStack-5B maintains robust general multimodal and spatial-temporal reasoning capabilities, demonstrating no catastrophic forgetting.}
    \label{tab:general_eval}
    \vspace{-4mm}
\end{table}

\subsection{Ablation Study}
\label{sec:ablation}
\paragraph{VGGT Layer Selection Ablation.}
Our selection mirrors VGGT's default indices \{4, 11, 17, 23\}, but with one adjustment: we excluded layer 4 due to poor performance in preliminary testing (insufficient network depth). This set provides a representative spread of shallow, middle, and deep features. \cref{tab:layer_trend} shows that replacing layer L23 with L21 or L22 (either alone or via multi-layer fusion with L11 and L17, denoted by ``+") yields no significant performance changes. This confirms that broad sampling across network depth is more critical than specific layer indices.

\begin{table}[t]
      \vspace{-3mm}
      \centering
      \small
      \resizebox{0.95\linewidth}{!}{
      \begin{tabular}{l|ccc|ccc}
          \toprule
          \textbf{Metric} & \textbf{L21} & \textbf{L22} & \textbf{L23} & +\textbf{L21} & +\textbf{L22} & +\textbf{L23} \\
          \midrule
          Low-Level & 63.54 & 64.87 & 64.33 & 65.89 & 65.45 & 64.44 \\
          High-Level & 65.57 & 66.51 & 66.36 & 65.95 & 66.78 & 67.52 \\
          \bottomrule
      \end{tabular}}
      \vspace{-2mm}
      \caption{\textbf{Layer Selection Ablation.} Performance comparison of extracting geometry features from different deep VGGT layers (L21, L22, L23) and their multi-layer combinations. }
      \vspace{-0mm}
      \label{tab:layer_trend}
  \end{table}

\paragraph{Geometry-Language Fusion Order Ablation.}
We analyze the order of layer-wise geometry-language feature fusion in~\cref{tab:ablation_fusion_order}. SpatialStack uses a progressive hierarchical mapping (L11 stands for layer 11): Geo-L11 $\rightarrow$ LLM-L0, Geo-L17 $\rightarrow$ LLM-L1, and Geo-L23 $\rightarrow$ LLM-L2. This is compared against a Reverse configuration (Geo-L11 $\rightarrow$ LLM-L2, Geo-L17 $\rightarrow$ LLM-L1, Geo-L23 $\rightarrow$ LLM-L0). 
SpatialStack outperforms both of the Reverse baseline and Vision Fusion baseline in 3 out of 4 benchmarks and achieves a higher overall score, confirming our hierarchical alignment is optimal for preserving spatial information.

\begin{table}[t]
    \centering
    \small
    \resizebox{1.0\linewidth}{!}{
        \renewcommand{\arraystretch}{0.85}
        \setlength{\tabcolsep}{4pt}
        \begin{tabular}{l|cccc|c}
            \toprule
            Methods & 
            \begin{tabular}[c]{@{}c@{}}VSI\\Bench\end{tabular} & 
            \begin{tabular}[c]{@{}c@{}}SPAR\\Bench\end{tabular} & 
            \begin{tabular}[c]{@{}c@{}}BLINK\\Spatial\end{tabular} & 
            \begin{tabular}[c]{@{}c@{}}CV\\Bench\end{tabular} & 
            \begin{tabular}[c]{@{}c@{}}Overall\end{tabular} \\ 
            \midrule
            Qwen3.5
            & 64.76
            & 68.75
            & 56.10
            & 84.49
            & 68.52 \\
            Vision Fusion
            & 64.27
            & 69.68
            & \textbf{56.45}
            & 83.11
            & 68.38 \\
            SpatialStack (Reverse)
            & 67.22
            & \textbf{71.97}
            & 50.08
            & 84.82
            & 68.52 \\
            SpatialStack (final)
            & \textbf{67.52}
            & 71.39
            & 52.12
            & \textbf{85.53}
            & \textbf{69.14} \\
            \bottomrule
        \end{tabular}
    }
    \vspace{-2mm}
    \caption{\textbf{Geometry-Language Fusion Order Ablation.} Comparison of our progressive hierarchical alignment against a reverse fusion strategy and baseline models.}
    \label{tab:ablation_fusion_order}
    \vspace{-3mm}
\end{table}

\section{Conclusion}
\label{sec:conclusion}
We introduced SpatialStack, a hierarchical fusion framework bridging the gap between vision, geometry, and language for robust 3D spatial reasoning. Our layer-wise analysis reveals a key correspondence: shallow geometry layers preserve fine-grained spatial details, while deeper layers capture global semantic context. We find that naive multi-layer geometry-vision fusion creates a structural bottleneck, leading to feature interference rather than synergy. By progressively aligning multi-level geometric features with the LLM decoder, SpatialStack preserves both local precision and high-level relational semantics. Extensive evaluations across multiple 3D benchmarks show that our approach achieves state-of-the-art performance among open-source models, exhibiting strong zero-shot generalization without compromising general multimodal capabilities. SpatialStack establishes a new paradigm for vision-language-geometry integration, paving the way for AI systems that truly understand and act within the physical 3D world.

\clearpage
{
    \small
    \bibliographystyle{ieeenat_fullname}
    \bibliography{main}
}

\clearpage
\setcounter{section}{0}
\setcounter{figure}{0}
\setcounter{table}{0}
\maketitlesupplementary

\renewcommand\thesection{\Alph{section}} 
\renewcommand\thesubsection{\thesection.\arabic{subsection}} 
\renewcommand\thefigure{\Alph{figure}} 
\renewcommand\thetable{\Alph{table}} 

In this supplementary material, we provide comprehensive implementation details and additional experimental results for \textit{SpatialStack}. The content is organized as follows:
\begin{itemize}
    \item \cref{sec:spatialstack} elaborates on the detailed architectural components, including the geometry token extraction pipeline and the masked additive fusion mechanism.
    \item \cref{sec:dataset_details} describes the composition and statistics of our training dataset mixture.
    \item \cref{sec:training} describes the training details, including input processing and specific training configurations.
    \item \cref{sec:evaluation} provides the detailed evaluation protocols, including the specific benchmarks and metrics.
    \item \cref{sec:more-results} presents additional baseline comparisons on zero-shot spatial reasoning in CV-Bench.
    \item \cref{sec:more-visualizations} offers qualitative visualizations contrasting the feature responses of geometry and vision encoders.
\end{itemize}

\section{Architecture Details}
\label{sec:spatialstack}
To enable both fine-grained and global spatial reasoning, our architecture integrates multi-level geometric cues
extracted from VGGT~\cite{wang2025vggt} into the VLM. The overall pipeline consists of three stages: geometry token
extraction and spatial alignment (\cref{sec:geo_tokens}); geometry merging and projection into the language feature space
(\cref{sec:geo_merge}); and masked additive fusion that injects geometry exclusively into the visual-token slice of the
decoder state (\cref{sec:geo_fusion}). The following subsections describe each component in detail.

\subsection{Geometry Token Extraction and Preprocessing}
\label{sec:geo_tokens}

We first outline the end-to-end flow of geometry token extraction and alignment before detailing the reshaping and
reordering process. At both training and inference time, images are processed by the vision encoder of the chosen base
model (Qwen2.5-VL~\cite{bai2025qwen2.5vl} or Qwen3.5~\cite{qwen35blog}) to generate visual features. For Qwen2.5-VL, this
encoding procedure consists of patch embedding, window based attention, and hierarchical patch merging, and produces a
sequence of merged vision tokens; for Qwen3.5, the stock vision encoder produces image embeddings that are inserted into
the multimodal sequence. In parallel, VGGT (frozen, evaluation mode) emits geometry tokens or layer-wise geometry
features from selected internal aggregator layers. These geometry features are subsequently reshaped and reordered, when
needed, to match the layout of the visual tokens before fusion, ensuring spatial consistency.

\paragraph{Token Structuring.}
VGGT produces a sequence of tokens at multiple internal aggregator layers. Each output contains three types of tokens: a \textit{camera token} encoding global viewpoint information; several \textit{register tokens} acting as global latent slots; and a sequence of \textit{patch tokens} representing per-patch geometric features.

Let \(h_{\text{patch}} = H/p\) and \(w_{\text{patch}} = W/p\) denote the spatial resolution of the VGGT patch tokens. The patch tokens are originally arranged in a flat row-major sequence of length \(h_{\text{patch}} \times w_{\text{patch}}\). To align their traversal order with the vision encoder after the spatial merger step, we partition the spatial grid into windows of size \(s \times s\), where \(s = \texttt{spatial\_merge\_size}\) (default \(s = 2\)):

\begin{equation}
(h_{\text{patch}},\, w_{\text{patch}})
\;\rightarrow\;
\bigl(\tfrac{h_{\text{patch}}}{s},\, s,\, \tfrac{w_{\text{patch}}}{s},\, s\bigr),
\label{eq:reshape}
\end{equation}

and apply a permutation that moves window indices ahead of within-window positions:
\begin{equation}
\bigl(\tfrac{h_{\text{patch}}}{s},\, s,\, \tfrac{w_{\text{patch}}}{s},\, s\bigr)
\;\rightarrow\;
\bigl(\tfrac{h_{\text{patch}}}{s},\, \tfrac{w_{\text{patch}}}{s},\, s,\, s\bigr).
\label{eq:permute}
\end{equation}

Finally, the reordered grid is flattened back into a 1D sequence:
\begin{equation}
\bigl(\tfrac{h_{\text{patch}}}{s},\, \tfrac{w_{\text{patch}}}{s},\, s,\, s\bigr)
\;\rightarrow\;
h_{\text{patch}} \cdot w_{\text{patch}}.
\label{eq:flatten}
\end{equation}

This reordering preserves the total number of tokens while changing their traversal order: tokens are enumerated window-by-window rather than row-by-row. As a result, consecutive groups of \(s^2\) geometry tokens correspond to the same spatial region grouped by one merged
  visual token, ensuring spatial alignment prior to fusion.

\subsection{Geometry-to-Language Projection}
\label{sec:geo_merge}
After reordering the geometry patch tokens to match the traversal order of the merged vision tokens (Sec.~\ref{sec:geo_tokens}), we obtain a 1D sequence
\begin{equation}
\mathbf{Z} \in \mathbb{R}^{(h_{\text{patch}} \cdot w_{\text{patch}}) \times D_{\text{geo}}},
\label{eq:z_init}
\end{equation}
where \(h_{\text{patch}} \cdot w_{\text{patch}}\) denotes the total number of spatial tokens and \(D_{\text{g}}\) is the geometry feature dimension.

\noindent\textbf{Normalization.}
Following the design of Qwen2.5-VL~\cite{bai2025qwen2.5vl}, token-wise RMS normalization~\cite{zhang2019root} is first applied:
\begin{equation}
\mathbf{Z}_{\text{norm}} = \mathrm{RMSNorm}(\mathbf{Z}).
\end{equation}

\noindent\textbf{Window-wise merging.}
The normalized tokens are grouped into non-overlapping spatial windows of size \(s \times s\). 
Each window is flattened and concatenated along the channel dimension, producing a 1D sequence of merged geometry tokens:
\begin{equation}
\tilde{\mathbf{Z}} \in
\mathbb{R}^{\left(\frac{h_{\text{patch}}}{s} \cdot \frac{w_{\text{patch}}}{s}\right) \times (s^2 D_{\text{geo}})}.
\label{eq:geo_window_merge}
\end{equation}

\noindent\textbf{Projection to language space.}
Each flattened window token is projected to the language decoder dimension by a two-layer MLP:
\begin{equation}
\mathbf{G} = W_2\,\sigma(W_1 \tilde{\mathbf{Z}} + b_1) + b_2,
\label{eq:geo_projector}
\end{equation}
where \(W_1 \in \mathbb{R}^{D_{\text{mlp}} \times (s^2D_{\text{geo}})}\),
\(b_1 \in \mathbb{R}^{D_{\text{mlp}}}\),
\(W_2 \in \mathbb{R}^{D_{\text{lang}} \times D_{\text{mlp}}}\),
and \(b_2 \in \mathbb{R}^{D_{\text{lang}}}\).
The projected geometry representation has shape
\begin{equation}
\mathbf{G} \in
\mathbb{R}^{\left(\frac{h_{\text{patch}}}{s} \cdot \frac{w_{\text{patch}}}{s}\right) \times D_{\text{lang}}}.
\label{eq:geo_final}
\end{equation}

\subsection{Additive Fusion via Vision-Token Mask}
  \label{sec:geo_fusion}

  Let \(H_l \in \mathbb{R}^{N_{\text{tot}} \times D_{\text{lang}}}\) denote the decoder
  hidden states at layer \(l\), where \(N_{\text{tot}}\) is the token sequence length
  (including system prompt, instruction text, vision tokens, and autoregressive text), and
  \(D_{\text{lang}}\) is the decoder hidden dimension.

  The projected geometry features from Sec.~\ref{sec:geo_merge} are
  \(\mathbf{G}_l \in \mathbb{R}^{N_p \times D_{\text{lang}}}\), where
  \(N_p\) denotes the number of vision tokens participating in fusion
  (in the default setting without camera tokens and assuming \(h_{\text{patch}}\) and \(w_{\text{patch}}\) are divisible by
  \(s\), \(N_p = \tfrac{h_{\text{patch}}}{s}\cdot\tfrac{w_{\text{patch}}}{s}\)).
  To locate the visual portion of the sequence, we define a binary mask
  \(M_{\text{vis}} \in \{0,1\}^{N_{\text{tot}}}\), where \(M_{\text{vis}}[i]=1\) if and only if
  position \(i\) corresponds to a visual token.

  Additive fusion updates only the masked positions:
  \begin{equation}
  \mathbf{H}_l \leftarrow \mathbf{H}_l + \mathrm{scatter}\big(\mathbf{G}_l,\ M_{\text{vis}}\big),
  \label{eq:add_fusion_mask}
  \end{equation}
  where \(\mathrm{scatter}(\mathbf{G}_l, M_{\text{vis}})\) distributes rows of \(\mathbf{G}_l\)
  sequentially to locations where \(M_{\text{vis}}=1\) and inserts zeros elsewhere.

  Equivalently, for each token index \(i\),
  \begin{equation}
  \mathbf{H}_l[i] \leftarrow
  \begin{cases}
  \mathbf{H}_l[i] + \mathbf{G}_l[k], & \text{if } M_{\text{vis}}[i]=1,\\[4pt]
  \mathbf{H}_l[i], & \text{if } M_{\text{vis}}[i]=0,
  \end{cases}
  \label{eq:piecewise}
  \end{equation}
  where \(k\) enumerates the \(N_p\) masked positions.

  Thus, geometry information is injected exclusively into the vision-token slice of the
  decoder state, while non-vision tokens (e.g., system prompt and text tokens) remain
  unchanged. During autoregressive generation, this fusion is applied at the initial
  prefill step, after which standard decoding proceeds with the updated hidden states.

\section{Dataset Details}
\label{sec:dataset_details}

\begin{figure*}[t]
    \centering
    \includegraphics[width=0.92\linewidth]{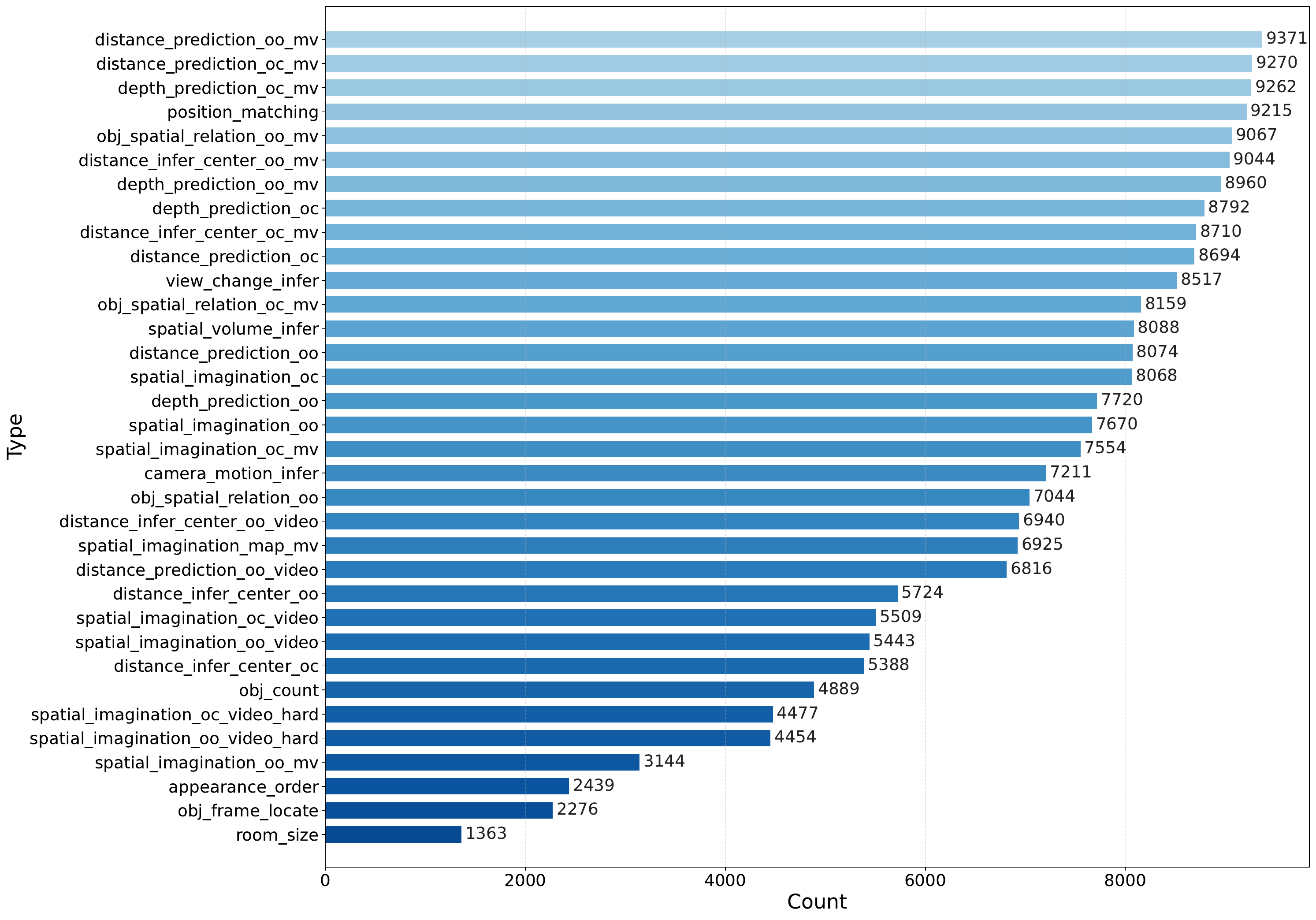}
    \caption{\textbf{Task-type distribution of the sampled SPAR subset.} The bar chart reports the counts of all 33 spatial task types after randomly sampling 60\% of SPAR-234k for training.}
    \label{fig:spar_tasks}
\end{figure*}

\begin{figure*}[t]
    \centering
    \includegraphics[width=0.92\linewidth]{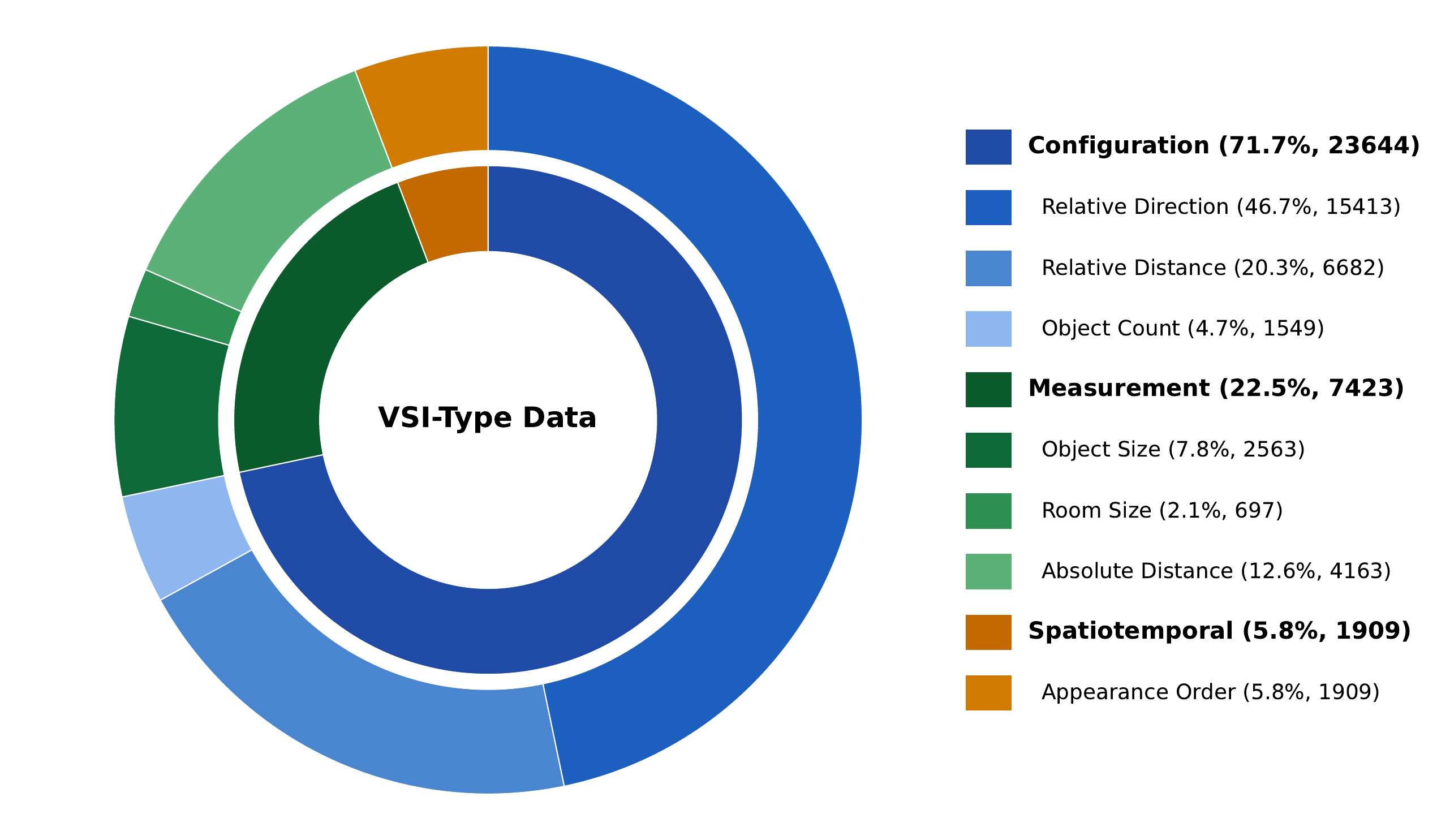}
    \caption{\textbf{Task-type distribution of the seven tasks in the VSI-Bench setting.} The pie chart summarizes the combined composition from VLM3R-ScanNet and the sampled Appearance-Order subset from VSI-590K, which are merged for unified reporting.}
    \label{fig:vsi_tasks}
\end{figure*}

We construct a balanced dataset of approximately \(200\mathrm{k}\) samples, blending spatial expertise with general instruction-following capabilities to facilitate efficient experimentation. Specifically, we sample subsets from \textbf{SPAR-234k} and \textbf{LLaVA-Hound-64k} (both from \textit{VG-LLM}~\cite{zheng2025learning}), as well as the \textbf{ScanNet split} of the \textit{VLM-3R} dataset~\cite{fan2026vlm3r}, which provides explicit spatial supervision. To enhance perception of object sequences, we additionally include approximately \(2\mathrm{k}\) appearance-order instances from the \textbf{VSI-590k}~\cite{yang2025cambrian} collection. As summarized in \cref{tab:dataset_summary}, this composition ensures broad task coverage suitable for controlled architectural ablations.

\subsection{Spatial Instruction-Following Data}

\noindent\textbf{SPAR (Spatial Perception and Reasoning).}
SPAR~\cite{zhang2025flatland} is a large-scale vision–language dataset designed for \textit{spatial perception and reasoning} in complex indoor scenes, featuring diverse question–answer pairs across 33 spatial task types spanning low-level perception to high-level reasoning, and covering single-view, multi-view, and video formats. We build upon the publicly released \textbf{SPAR-234k} subset introduced in \cite{zheng2025learning}; the detailed task-type distribution of our sampled training set is illustrated in \cref{fig:spar_tasks}.

\noindent\textbf{VLM-3R.}
VLM-3R is a spatial QA construction framework based on open-source 3D datasets with geometry, semantic labels, and instance-level annotations, including ScanNet~\cite{dai2017scannet}, ScanNet++~\cite{yeshwanth2023scannet++}, and ARKitScenes~\cite{baruch2021arkitscenes}. We use only the ScanNet split, which provides six spatial QA task types: Object Counting, Relative Distance, Relative Direction, Object Size, Absolute Distance, and Room Size. This split does not include Route Planning or Appearance Order tasks.

\noindent\textbf{VSI-590K.}
VSI-590k is a large-scale spatial instruction-tuning dataset consisting of 590k QA examples from real and simulated indoor environments across 12 task types. For training, we extract a 2k subset corresponding to the appearance-order task derived specifically from the ScanNet portion of VSI-590k, which supplements the absence of appearance-order supervision in the VLM-3R ScanNet split.

We refer to this combined compilation of spatial tasks as \textbf{VSI-Type Data}. As visualized in \cref{fig:vsi_tasks}, these seven tasks are categorized into three major groups: Configuration, Measurement, and Spatiotemporal, following the taxonomy in the VSI-Bench setting.

\subsection{General Video Instruction-Following Data}

\noindent\textbf{LLaVA-Hound.}
LLaVA-Hound~\cite{zhang2025direct} is a dataset for video captioning, instruction tuning, and preference alignment,
curated from 900k videos sourced from WebVid, VIDAL, and ActivityNet. High-quality captions
are produced using GPT-4V from uniformly sampled frames, followed by 240k instruction–answer
pairs generated using ChatGPT and 17k preference pairs for Direct Preference Optimization.
We use the 64k LLaVA-Hound subset released in VG-LLM, from which 60 percent is sampled to
retain general instruction-following and object-grounded reasoning capability while keeping
the training scale computationally manageable.

\begin{table}[t]
\centering
\footnotesize
\setlength{\tabcolsep}{4pt}
\begin{tabular}{lcc}
\toprule
\textbf{Dataset} & \textbf{Raw} & \textbf{Train Subset} \\
\midrule
SPAR 234k            & 234k (66.3\%) & 140k (66.4\%) \\
LLaVA-Hound 64k      & 63.8k (18.0\%) & 38.3k (18.1\%) \\
VLM3R-ScanNet        & 51.8k (14.6\%) & 31.1k (14.7\%) \\
VSI App-Order        & 3.8k (1.1\%)   & 1.9k (0.9\%) \\
\midrule
\textbf{Total}       & 353k (100\%)   & 212k (100\%) \\
\bottomrule
\end{tabular}
\caption{
\textbf{Dataset scales and sampled subsets used in our \(\sim 200\mathrm{k}\) training mixture.}
We sample \(60\%\) from SPAR-234k, LLaVA-Hound-64k~\cite{zheng2025learning}, and the ScanNet
split of VLM-3R~\cite{fan2026vlm3r}, and add \(\sim 2\mathrm{k}\) appearance-order instances from
VSI-590k~\cite{yang2025cambrian} to compensate for the missing ordering supervision.
Percentages indicate each dataset’s contribution to the final mixture.
}
\label{tab:dataset_summary}
\end{table}

\section{Training Details}
\label{sec:training}

This section details the implementation of \textit{SpatialStack}, focusing on (1) input processing and (2) {training settings.
The model is trained via large-scale geometry-aware instruction tuning, where only the language tower and geometry-merger modules are updated, while the vision tower and VGGT remain frozen. All experiments are conducted on 32 NVIDIA A100 GPUs (80GB).

\subsection{Input Processing}
\label{sec:input-processing}

Videos are first decomposed into individual frame images before entering the multimodal pipeline.
A single video token in the prompt is expanded into \(K\) consecutive image tokens.
For a clip of duration \(T_{\text{sec}}\) containing \(F\) total frames, we uniformly sample
\(K = \mathrm{clip}\!\left(\mathrm{round}\!\left(T_{\text{sec}} / \Delta \right), K_{\min}, K_{\max}\right)\)
frame indices from \([0, F-1]\), where \(\Delta\) denotes the temporal sampling interval.

Each frame (and standalone image) undergoes a unified visual preprocessing pipeline.
For SPAR-style training samples, optional task-specific marking is first applied on the original-resolution image: task cues such as points or bounding boxes are drawn according to the provided annotation metadata before any resizing.
Transparency is then composited onto a white background and the image is converted to RGB.

Next, we resize the image while preserving aspect ratio to a target size of 518 pixels. In the default crop-based
setting, one side is resized to 518 pixels and the other side is scaled proportionally, with center cropping applied when needed. We then apply, when necessary, patch-aligned spatial trimming so that the final height and width satisfy \(H \bmod (p \cdot m) = 0\) and \(W \bmod (p \cdot m) = 0\), ensuring that the resolution becomes an integer multiple of the effective patch unit \(p \cdot m\) (e.g., \(14 \cdot 2 = 28\)). This alignment is required because the merge stage groups \(m \times m\) adjacent patches into a single token.

Finally, the resized image is used to construct inputs for both the vision encoder and the geometry encoder (VGGT), with
additional patch/merge alignment applied where needed to maintain spatial consistency between the two branches.

\subsection{Training Settings}
\label{sec:train-settings}
We train SpatialStack using \texttt{torchrun} with DeepSpeed ZeRO-2. Optimization uses AdamW with cosine decay scheduling and warmup. bfloat16 precision is employed for training efficiency and numerical robustness.~\cref{tab:training_hparams} summarizes the configuration.                                                                                            
                                                                                     
  \begin{table}[h]
  \centering
  \footnotesize
  \setlength{\tabcolsep}{5pt}
  \renewcommand{\arraystretch}{1.05}
  \begin{tabular}{ll}
  \toprule
  \textbf{Category} & \textbf{Setting} \\
  \midrule
  Base model & Qwen2.5-VL-3B or Qwen3.5-4B \\
  Geometry encoder & VGGT-1B (frozen) \\
  Fusion strategy & SpatialStack (multi-depth) \\
  Trainable modules & Language tower + fusion modules \\
  \cmidrule(lr){1-2}
  Precision & bfloat16 \\
  Optimizer & AdamW (wd=0.01) \\
  Learning rate & $1{\times}10^{-5}$ \\
  Scheduler & Cosine decay, warmup 3\% \\
  Epochs & 1 \\
  \cmidrule(lr){1-2}
  Batch size & effective 64 \\
  Sequence length & 12{,}800 tokens \\
  Frames per video & 4–8 \\
  Pixels/sample & $16\!\cdot\!28^2$--$576\!\cdot\!28^2$ \\
  \cmidrule(lr){1-2}
  Distributed & torchrun + DeepSpeed ZeRO-2 \\
  Checkpoint save interval & every 1000 steps \\
  Logging & every 10 steps \\
  Hardware & 32$\times$A100 GPUs (80GB) \\
  \bottomrule
  \end{tabular}
  \caption{\textbf{Training hyperparameters for SpatialStack.}
  Geometry-aware instruction tuning is performed on Qwen2.5-VL-3B or Qwen3.5-4B with VGGT-1B using the proposed
  SpatialStack fusion.
  The language tower and fusion modules are trainable, while the geometry encoder remains frozen.
  Training uses AdamW (bfloat16, cosine schedule) with an effective batch size of 64 under ZeRO-2 parallelism.}
  \label{tab:training_hparams}
  \end{table}

\begin{figure*}[t]
    \centering
    \includegraphics[width=1.0\textwidth]{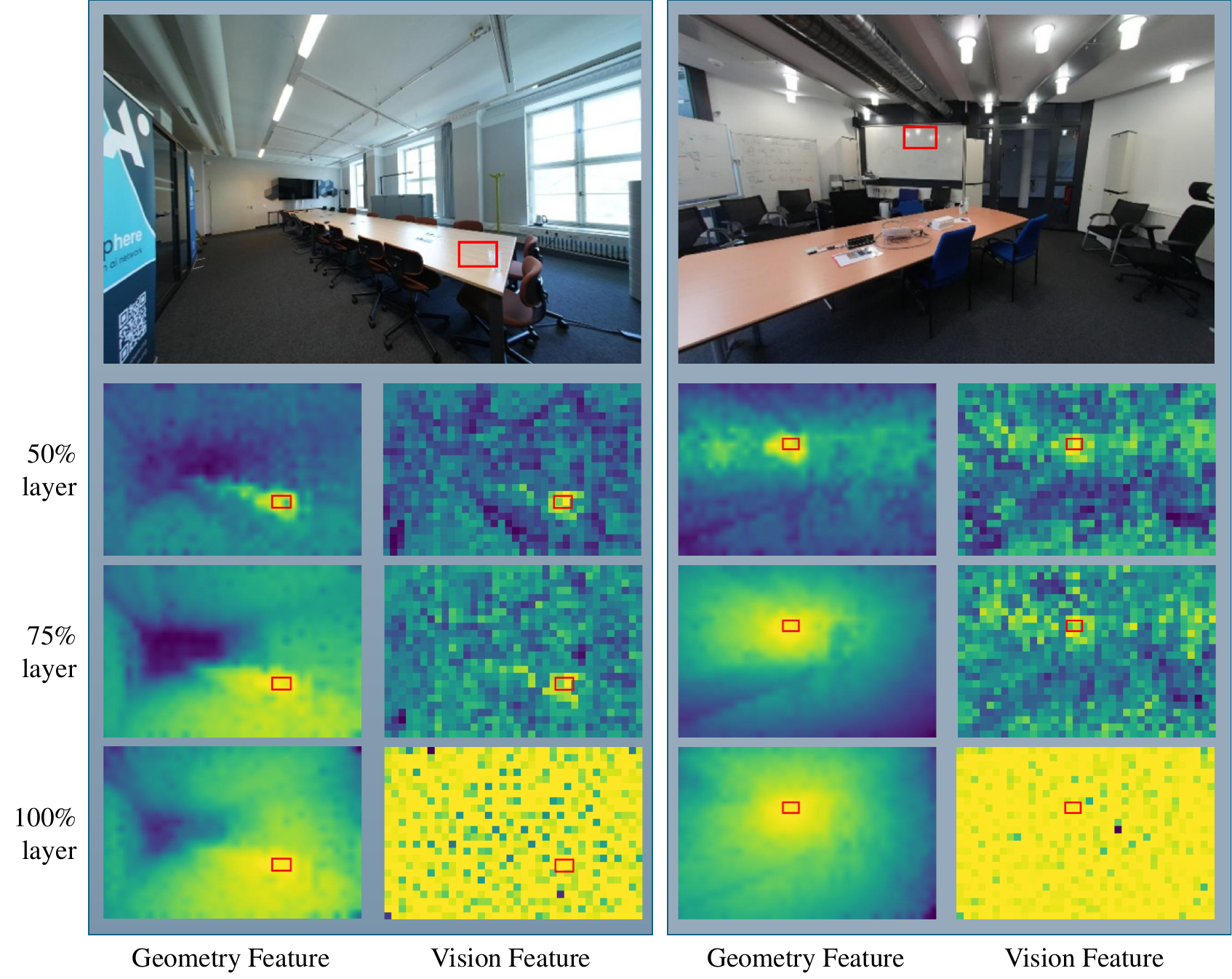}
    \vspace{-3mm}
    \caption{
    \textbf{ROI similarity comparison between geometry and vision features across encoder depths.}
    For two indoor scenes, the top row shows the RGB image with the ROI marked in red. The lower rows display similarity maps (brighter means more similar) at \(50\%\), \(75\%\), and \(100\%\) depths of the geometry encoder (left) and the vision encoder (right). Geometry features preserve meaningful spatial structure, while vision features are noisy and become nearly uniform at deeper layers.
    }
    \label{fig:geo-vs-vis-layers}
    \vspace{-2mm}
\end{figure*}

\section{Evaluation Details}
  \label{sec:evaluation}

  Our evaluation pipeline closely follows established protocols to ensure fair comparison. Specifically, we adopt the data
  preprocessing methodology from \textit{VG-LLM}~\cite{zheng2025learning} and adhere to the standard evaluation parameter
  settings defined in \textit{VSI-Bench}~\cite{yang2025thinking}.

  \noindent\textbf{Implementation Details.}
  Visual inputs (single images, image lists, or videos) are first decomposed into sampled frames with a capped count \(K\),
  using uniform frame sampling in the evaluation pipeline.
  Following the preprocessing pipeline of \cite{zheng2025learning}, geometry-aware evaluation uses a 518-pixel image
  preprocessing step.
  To ensure compatibility with our token merging mechanism, patch/merge alignment is enforced when required so that the
  patch grid dimensions are divisible by the merge factor \(m\):
  \begin{equation}
  (W/p) \bmod m = 0 \quad \text{and} \quad (H/p) \bmod m = 0,
  \end{equation}
  where \(p\) denotes the patch size.
  When geometry is enabled, the geometry encoder inputs are constructed from the same visual content as the vision branch
  to maintain spatial correspondence.

  Geometry tokens are computed once per sample in evaluation mode. Geometry fusion is injected at predefined decoder layers
  after self-attention and MLP execution, replacing the vision-aligned slice before decoding continues.

  Decoding adopts greedy generation by default (\(\texttt{temperature}=0\), \(\texttt{num\_beams}=1\)) with task-specific
  generation limits unless specified otherwise. Key/value caching is enabled for efficiency, and outputs are trimmed to
  remove the prompt prefix before evaluation. All benchmark results in the main paper are produced under this evaluation
  configuration.
  
\section{More Results}
\label{sec:more-results}

We evaluate zero-shot spatial reasoning on CV-Bench in~\cref{tab:cvbench}. Our method consistently outperforms both SpatialRGPT~\cite{cheng2024spatialrgpt} and Spatialbot~\cite{cai2025spatialbot} across all metrics. Note that SpatialVLM~\cite{chen2024spatialvlm} is excluded as its code is unavailable.

  \begin{table}[h]
      \vspace{-2mm}
      \centering
      \small
      \setlength{\tabcolsep}{3pt}
      \resizebox{1.0\linewidth}{!}{
      \begin{tabular}{l|cccc|c}
          \toprule
          \textbf{Method} & \textbf{Count} & \textbf{Relation} & \textbf{Depth} & \textbf{Distance} & \textbf{Overall} \\
          \midrule
          SpatialRGPT & 60.4 & 78.9 & 80.0 & 71.3 & 72.7 \\
          Spatialbot & 61.4 & 73.1 & 76.5 & 61.0 & 68.0 \\
          \textbf{Ours} & \textbf{69.0} & \textbf{92.5} & \textbf{93.7} & \textbf{90.7} & \textbf{86.5} \\
          \bottomrule
      \end{tabular}
      }
      \vspace{-10pt}
      \caption{\textbf{Additional Baseline Comparison on CV-Bench.}}
      \label{tab:cvbench}
      \vspace{-4mm}
  \end{table}

\section{More Visualizations}
\label{sec:more-visualizations}
\subsection{Geometry vs.\ Vision Feature Responses}
\label{sec:geo-vs-vis-response}

To analyze the difference between geometry and vision representations, we visualize ROI-based similarity maps derived from features at different encoder depths, as shown in \cref{fig:geo-vs-vis-layers}. For each scene, a red box marks a region of interest (ROI) in the RGB image. We compute patch-wise similarity between this ROI and all other spatial locations using features extracted at \(50\%\), \(75\%\), and \(100\%\) depth of the geometry encoder and compare them with features from the native vision encoder at corresponding relative depths.

Here, the percentages refer to proportional positions within the encoder stack rather than absolute layer indices. For example, the geometry encoder contains 24 layers, so \(50\%\), \(75\%\), and \(100\%\) depths correspond to layers 12, 18, and 24. The vision encoder contains 32 layers, where the same relative depths map to layers 16, 24, and 32. This proportional alignment allows a fair comparison between encoders with different depths.

The similarity maps reveal a consistent trend: shallow geometry layers preserve fine-grained spatial distinctions and clear geometric boundaries, whereas deeper geometry layers become increasingly homogeneous, causing many regions to appear similar despite different physical geometry. In contrast, similarity maps from the native visual encoder are noisy and spatially fragmented across depths, and at the deepest layers they collapse into nearly uniform responses without meaningful spatial differentiation.

These results demonstrate that internal visual features alone lack explicit spatial structure and are insufficient for reasoning about relative geometry. External geometry encoders provide structured spatial cues at different levels that are missing from the native visual pathway, motivating the use of multi-level geometry fusion in spatial reasoning.

\end{document}